\documentclass{article}

\PassOptionsToPackage{numbers,sort&compress}{natbib}
\usepackage[main,preprint]{neurips_2026}
\usepackage[utf8]{inputenc}
\usepackage[T1]{fontenc}
\usepackage{amsmath,amssymb,amsfonts}
\usepackage{booktabs}
\usepackage{graphicx}
\usepackage{microtype}
\usepackage{xcolor}
\usepackage{url}
\usepackage{hyperref}

\newcommand{\method}{PGP-Clinical-TimeKAN}
\newcommand{\R}{\mathbb{R}}
\newcommand{\vect}[1]{\boldsymbol{#1}}

\title{\method: Prior-Guided Joint Probabilistic Forecasting of Clinical Trajectories}

\author{%
Weizhi Nie \quad Rihao Chang \quad Weijie Wang \quad Yuting Su\\
Tianjin University
}

\begin{document}
\maketitle

\begin{abstract}
Clinical deterioration unfolds through coupled, partially observed trajectories, not a single diagnostic label. We introduce \method, a trajectory-first framework for joint probabilistic forecasting of multivariate physiology. It combines missingness-aware temporal encoders, a soft organ-system prior, patient-specific relations, nonlinear Kolmogorov--Arnold messages, and a low-rank multivariate Student-$t$ head. We evaluate 24-hour histories and six-hour forecasts on a frozen MIMIC-IV-derived cohort of 6,882 patients and 54,694 windows. Across five seeds and 13 models, \method{} obtains the second-lowest normalized MAE ($0.37727\pm0.00029$) and the lowest RMSE ($0.52656\pm0.00034$). It reduces MAE by 0.52\% relative to deterministic TimeKAN. For probabilistic forecasting, it reaches a marginal NLL of 0.66380 and a CRPS of 0.27301. Empirical coverage is 0.533, 0.831, and 0.958 for nominal 50\%, 80\%, and 95\% intervals. Removing relational structure causes the largest ablation loss. Increasing covariance rank improves joint likelihood but has little effect on point accuracy. A trajectory-derived risk score remains weaker than a dedicated GRU-D classifier (AUROC 0.603 versus 0.650), which limits the present clinical claim. Joint trajectory forecasting therefore provides an inspectable intermediate task, but accurate physiology forecasts alone do not ensure a calibrated event detector.
\end{abstract}

\section{Introduction}

Sepsis is life-threatening organ dysfunction caused by a dysregulated response to infection. Its operational definition requires suspected infection and an acute increase of at least two points in the Sequential Organ Failure Assessment (SOFA) score~\citep{singer2016sepsis3}. Early-warning studies therefore seek signs of impending sepsis or septic shock in routinely collected electronic health record (EHR) data~\citep{henry2015trewscore,nemati2018sepsis,reyna2020physionet}. Such systems can be clinically useful. Yet a direct history-to-label model says only whether an event is likely. It does not show which variables may change, when a threshold may be crossed, or whether several organ systems may deteriorate together.

Forecasting future physiology offers another route. From the available history, a model predicts the next several hours of vital signs and laboratory measurements. These trajectories can then inform clinical event estimates. The intermediate representation is inspectable: clinicians can compare central forecasts, uncertainty intervals, and sampled multiorgan courses. It also separates two questions often treated as one. Can future physiology be forecast reliably? If so, does that forecast support a calibrated warning?

Three properties make this task difficult. First, ICU measurements are irregular and informatively missing. Both the absence and age of a measurement can reflect clinical workflow~\citep{che2018grud}. Second, physiological variables evolve at different rates and interact across organ systems. Third, uncertainty spans variables and horizons. Mixing variables in a hidden representation does not create a joint predictive distribution. Independently sampled marginals can still produce physiologically discordant futures.

We address these challenges with Prior-Guided Probabilistic Clinical-TimeKAN (\method). The model builds a missingness-aware representation for each variable. Multi-scale TimeKAN blocks~\citep{huang2025timekan} extract variable-specific temporal patterns. Next, a soft organ-system graph is blended with patient-specific dynamic relations. Kolmogorov--Arnold network (KAN) functions~\citep{liu2024kan} transform messages along these relations. Thus, a cross-variable association can take a nonlinear form rather than collapse to an attention weight. A low-rank multivariate Student-$t$ distribution then models correlated uncertainty across variables and future times. Its coherent samples yield threshold-crossing and organ-deterioration probabilities.

\begin{figure}[t]
    \centering
    \includegraphics[width=\linewidth]{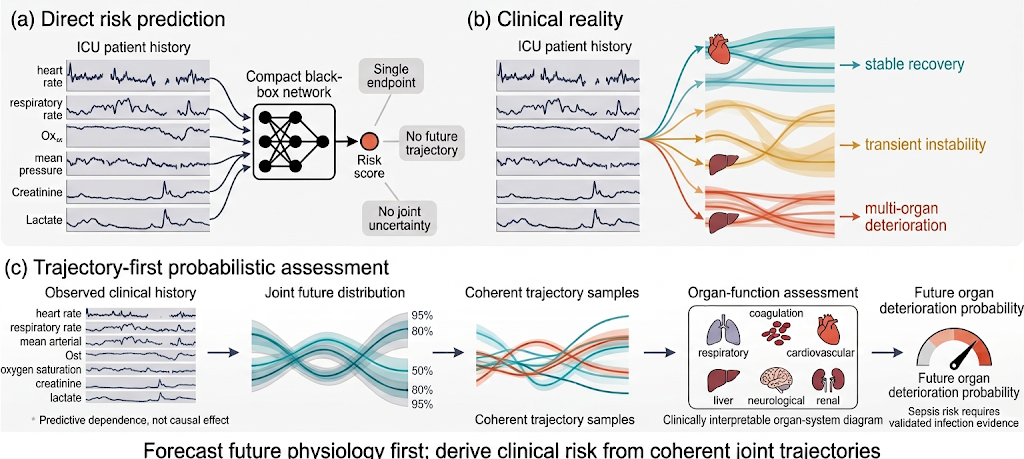}
    \caption{Motivation for trajectory-first clinical risk assessment. (a) Direct early-warning models compress the observed history into a scalar risk score without exposing the predicted physiological course. (b) The same clinical history can lead to multiple correlated futures involving different combinations of organ-system changes. (c) Our formulation predicts a joint distribution over future clinical trajectories and derives organ-deterioration probabilities from coherent multivariate samples. Learned dependencies are predictive rather than causal, and complete sepsis assessment additionally requires validated infection evidence.}
    \label{fig:motivation}
\end{figure}

Figure~\ref{fig:motivation} contrasts trajectory-first assessment with direct risk prediction. Direct models compress the observed history into a scalar score. Our formulation instead preserves several correlated futures and derives event probabilities from them.

The intended contributions are:
\begin{itemize}
    \item We formulate early clinical deterioration assessment as joint probabilistic forecasting of future multivariate physiology, preserving explicit trajectories between the observed history and the downstream risk estimate.
    \item We introduce a missingness-aware Clinical-TimeKAN architecture that combines multi-scale temporal encoding, a soft organ-system prior, patient-specific dynamic relations, and nonlinear KAN message functions.
    \item We develop a tractable low-rank multivariate Student-$t$ forecast over variables and horizons, enabling coherent trajectory sampling and explicit estimation of threshold-crossing and organ-deterioration events.
\end{itemize}
The frozen evaluation in Section~\ref{sec:experiments} shows competitive point accuracy, stronger probabilistic scores than the evaluated distributional baselines, and a clear limitation in downstream event discrimination.

\section{Related work}

\subsection{Early warning from clinical time series}

Early sepsis systems usually map an EHR history directly to a diagnostic endpoint. TREWScore used routine measurements to identify patients at risk of septic shock~\citep{henry2015trewscore}. Later work developed interpretable models for ICU sepsis prediction~\citep{nemati2018sepsis}. The PhysioNet/Computing in Cardiology Challenge 2019 formalized hourly early detection and introduced a utility function that balances timely alerts against false alarms~\citep{reyna2020physionet}. Prospective deployment of TREWS showed that retrospective discrimination, operational integration, and clinician response are separate issues~\citep{adams2022trews}. Systematic reviews also report wide variation in cohort construction, sepsis definitions, prediction horizons, validation, and reporting~\citep{fleuren2020systematic,moor2021systematic}.

Recent endpoint-oriented models add temporal, relational, or causal structure. TSCAN learns temporal--spatial attention over ICU variables for several prediction tasks~\citep{nie2024temporalspatial}. Causal-disentanglement models separate sepsis-related factors from other disease factors and confounders~\citep{li2025temporalspatial}. CISepsis instead uses causal diagrams, back-door adjustment, and instrumental-variable reasoning~\citep{li2024cisepsis}. These methods focus on robustness and interpretation at the diagnostic endpoint. Our goal is complementary and explicitly noncausal. We forecast a joint distribution over future physiology, then derive event risk from coherent trajectories. We do not claim to identify physiological effects. This distinction motivates an auditable intermediate target, especially given the heterogeneity of existing evaluations. We assess event risk only after evaluating trajectory accuracy and calibration.

\subsection{Irregular and missing clinical observations}

EHR observations are irregular and informative. Unstable patients may be measured more often, and a missing value is not equivalent to a normal one. GRU-D incorporates masks and elapsed time through learned decay~\citep{che2018grud}. BRITS jointly learns imputations and downstream representations with bidirectional recurrence~\citep{cao2018brits}. Continuous-time methods avoid a fixed grid. Latent ODEs model hidden dynamics between observations~\citep{rubanova2019latentode}; GRU-ODE-Bayes adds Bayesian updates at observation times~\citep{debrouwer2019gruode}; and neural controlled differential equations use the interpolated observation path as a control signal~\citep{kidger2020ncde}. These methods are well suited when observation times must be modeled directly. We retain an auditable hourly grid because the clinical endpoints require it. Measurement age and reliability remain explicit, and imputed future values never serve as targets.

\subsection{Multi-horizon time-series forecasting}

Multi-horizon forecasting now includes direct linear, convolutional, recurrent, and Transformer-based models. Informer introduced sparse attention and one-pass decoding for long sequences~\citep{zhou2021informer}. Autoformer combined progressive decomposition with autocorrelation~\citep{wu2021autoformer}. DLinear then showed that a carefully designed linear model remains a strong comparator~\citep{zeng2023dlinear}. PatchTST processes channel-independent temporal patches~\citep{nie2023patchtst}, whereas iTransformer treats variables as tokens to model cross-variate relations~\citep{liu2024itransformer}. TimesNet represents multi-period variation in two dimensions~\citep{wu2023timesnet}. Temporal Fusion Transformers combine static covariates, known inputs, gating, and quantile outputs~\citep{lim2021tft}. TimeKAN decomposes signals by frequency and applies multi-order KAN blocks to each component~\citep{huang2025timekan}. This range motivates our diverse baseline set. However, the standard versions do not jointly handle informative clinical missingness, organ-system priors, and correlated heavy-tailed uncertainty over the full forecast window.

\subsection{Graph-based multivariate dependencies and KANs}

Graph forecasters treat variables as nodes and learn cross-series interactions. Graph WaveNet augments a given graph with an adaptive dependency matrix~\citep{wu2019graphwavenet}. MTGNN learns directed relations and couples graph propagation with temporal convolutions~\citep{wu2020mtgnn}. StemGNN represents inter-series and temporal structure in the spectral domain~\citep{cao2020stemgnn}. Latent graphs can improve multivariate forecasts, but a fully learned graph may be unstable when clinical observations are sparse. A fixed expert graph can instead miss patient-specific relations. KANs replace scalar edge weights with learnable univariate functions~\citep{liu2024kan}; graph KAN variants extend this idea to relational learning~\citep{decarlo2024gkan}.

Structured priors also appear outside ICU forecasting. Prior knowledge has guided reinforcement-learning agents for thoracic-image diagnosis~\citep{nie2023deepreinforcement}. Structural causal models and back-door adjustment have been used for postoperative risk prediction after coronary artery bypass grafting~\citep{zhang2025causalinference}. These studies illustrate the broader value of domain structure, but they do not provide direct evidence for our time-series task. Our module blends a soft organ-system prior with patient-specific relations. KAN edge functions then model the nonlinear form of each message. We evaluate the resulting graph as a regularized predictive object, not as an identified causal graph.

\subsection{Joint probabilistic forecasting}

Proper scoring rules evaluate probabilistic forecasts more fully than point error alone~\citep{gneiting2007probabilistic}. DeepAR learns autoregressive distributions across related series~\citep{salinas2020deepar}, but factorized or sequential outputs may not preserve contemporaneous multivariate dependence. Low-rank Gaussian copula processes provide scalable joint forecasts for high-dimensional series~\citep{salinas2019deepvar}. More flexible alternatives include conditional normalizing flows, the autoregressive diffusion model TimeGrad~\citep{rasul2021timegrad}, and score-based models such as CSDI~\citep{tashiro2021csdi}. CSDI was designed for imputation but can be adapted to forecasting. Such generative models offer greater flexibility at the price of more complex training or iterative sampling. Our output head takes a simpler design point. A low-rank-plus-diagonal multivariate Student-$t$ distribution supports exact likelihoods on arbitrary observed subsets, heavy-tailed residuals, and coherent samples across variables and horizons.

\section{Problem formulation}
\label{sec:problem}

For patient $i$ at anchor $t$, let $\mathbf X_i\in\R^{L\times D}$ contain $D$ clinical variables over $L$ hourly bins. Let $\mathbf M_i\in\{0,1\}^{L\times D}$ be the observation mask, and let $\vect\Delta_i\in\R_{\geq 0}^{L\times D}$ record the time since each variable was last measured. Optional treatment histories and static covariates are denoted by $\mathbf A_i\in\R^{L\times D_a}$ and $\mathbf s_i\in\R^{D_s}$. The available history is
\begin{equation}
\mathcal H_{i,t}=\{\mathbf X_i,\mathbf M_i,\vect\Delta_i,\mathbf A_i,\mathbf s_i\}.
\end{equation}
The target is $\mathbf Y_i=\mathbf X_{i,t+1:t+H}\in\R^{H\times D}$ with future observation mask $\mathbf M_i^y$. Rather than estimating only $\mathbb E[\mathbf Y_i\mid\mathcal H_{i,t}]$, we learn
\begin{equation}
p_\theta(\mathbf Y_i\mid\mathcal H_{i,t}).
\end{equation}
Let $\mathbf y_i=\operatorname{vec}(\mathbf Y_i)\in\R^N$, where $N=HD$. This joint representation can encode dependencies between variables at a common horizon, within a variable across horizons, and across both variables and horizons.

\section{Data source and cohort construction}
\label{sec:data}

\begin{figure*}[t]
    \centering
    \includegraphics[width=\textwidth]{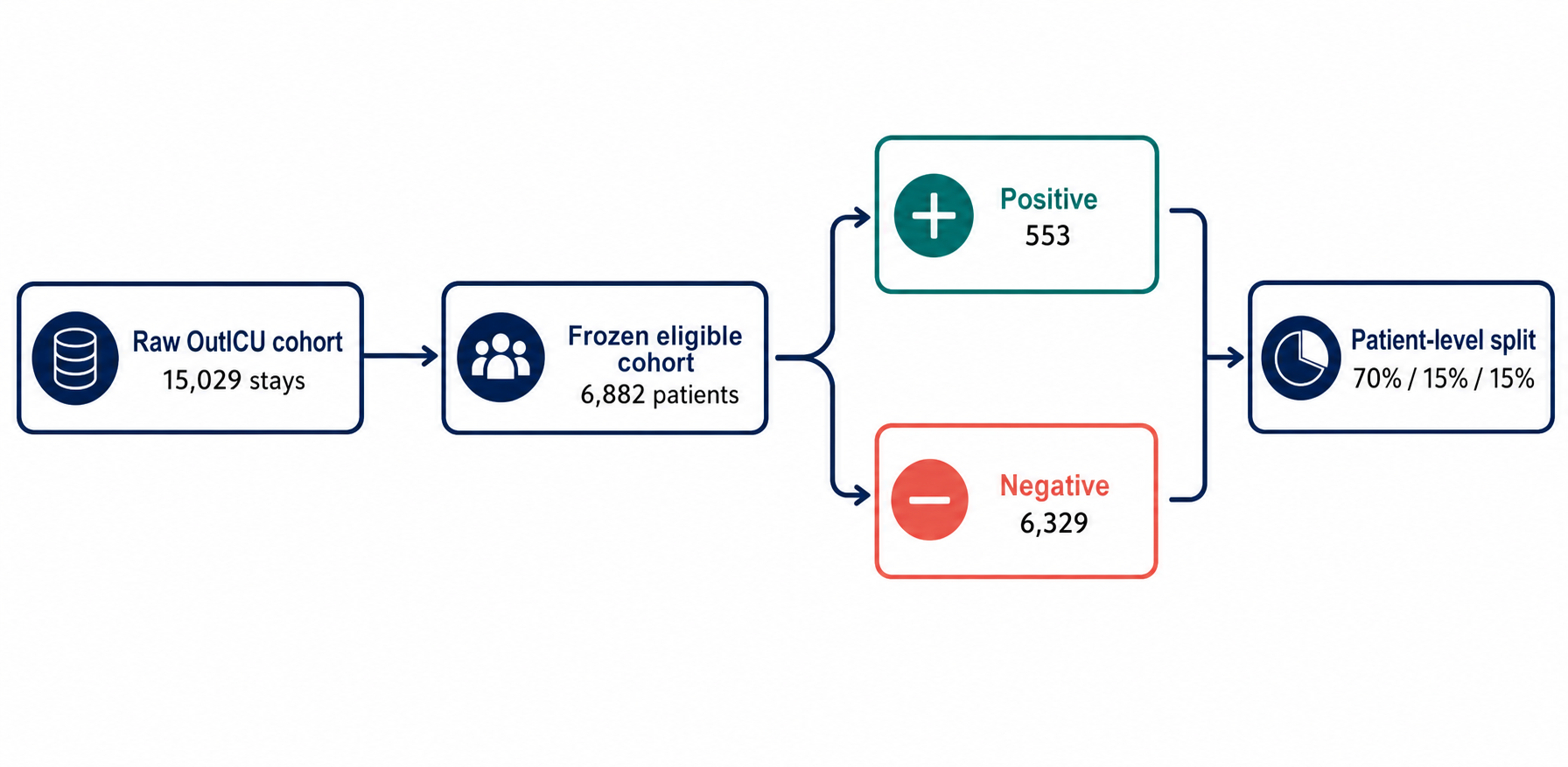}
    \caption{Frozen cohort construction. The raw 15,029-stay source view is reduced to 6,882 eligible patients, separated into 553 dataset-defined positive and 6,329 negative stays, and partitioned at the patient level. All stays and windows from one patient remain in a single split.}
    \label{fig:cohort-construction}
\end{figure*}

Figure~\ref{fig:cohort-construction} summarizes cohort screening, patient-level splitting, and leakage-controlled window construction. It also separates the dataset-defined positive label from a validated Sepsis-3 endpoint.

\subsection{Data source and frozen cohort}

We use a locally prepared MIMIC-IV time-series resource. MIMIC-IV is a deidentified, single-center critical-care EHR database~\citep{johnson2023mimiciv}. The maximal source view contains 15,029 ICU stays, 11,094,402 vital-sign events, and 3,327,232 laboratory or treatment events. Frozen eligibility and temporal-consistency rules leave 6,882 patients and the same number of stays. Of these stays, 553 are dataset-defined positives and 6,329 are negatives. Window construction yields 54,694 examples: 38,591 for training, 8,116 for validation, and 7,987 for testing.

The supplied derivative lacks independently auditable infection-evidence fields. We therefore call its source label a \emph{dataset-defined terminal-positive event}, not a validated Sepsis-3 endpoint. Masked multivariate trajectory quality is the primary outcome. The secondary risk analysis asks whether forecast samples predict the frozen dataset endpoint at the same anchors.

\subsection{Eligibility and leakage control}

We include adults ($\geq18$ years) with valid patient and ICU-stay identifiers. Admission and censoring times must be complete, with at least $L+H$ hours of usable follow-up. We exclude records with irreconcilable timestamps, unit conflicts, or label logic, and count every exclusion. Positive stays are censored at the label time. Negative stays are censored at ICU discharge or the last valid observation. For every positive window,
\begin{equation}
t_{\mathrm{input}}\leq t_{\mathrm{anchor}} < t_{\mathrm{label}},
\qquad t_{\mathrm{target}}\leq t_{\mathrm{label}}.
\end{equation}
All stays from a patient are assigned to a single split. Quality-control thresholds, imputation constants, scaling statistics, and calibration models are fitted on training data only.

\subsection{Windowing, variables, and preprocessing}

The primary setting uses $L=24$ hours of history and $H=6$ forecast hours. We aggregate data into one-hour bins and sample training anchors every six hours. Continuous variables use the within-hour median. Outcomes based on the worst hourly value, including some SOFA components, are computed in a separate derived view; they do not replace trajectory targets. The test set retains the natural event prevalence. Matched negative pseudo-anchors are used only for a secondary mechanism analysis, not for the primary risk estimate.

The six forecasting targets are heart rate, respiratory rate, oxygen saturation, mean arterial pressure, systolic pressure, and temperature. We use static metadata for descriptive and subgroup analyses, as well as a separate input ablation. The source derivative has no auditable time-varying treatment table. We therefore exclude treatment inputs from the reported main model and do not claim them as an evaluated contribution.

Forward filling uses only observations available by the anchor. If a variable has no earlier value in the window, it receives a training-set constant. We standardize continuous inputs with training-set medians and interquartile ranges. Future losses use observed targets only. Data audits cover mixed Celsius/Fahrenheit temperatures, implausible blood pressures, oxygenation units, renal and hepatic laboratory units, vasopressor dose equivalence, ventilation status, and sedation effects on GCS.

\section{Method}
\label{sec:method}

\begin{figure}[t]
    \centering
    \includegraphics[width=\linewidth]{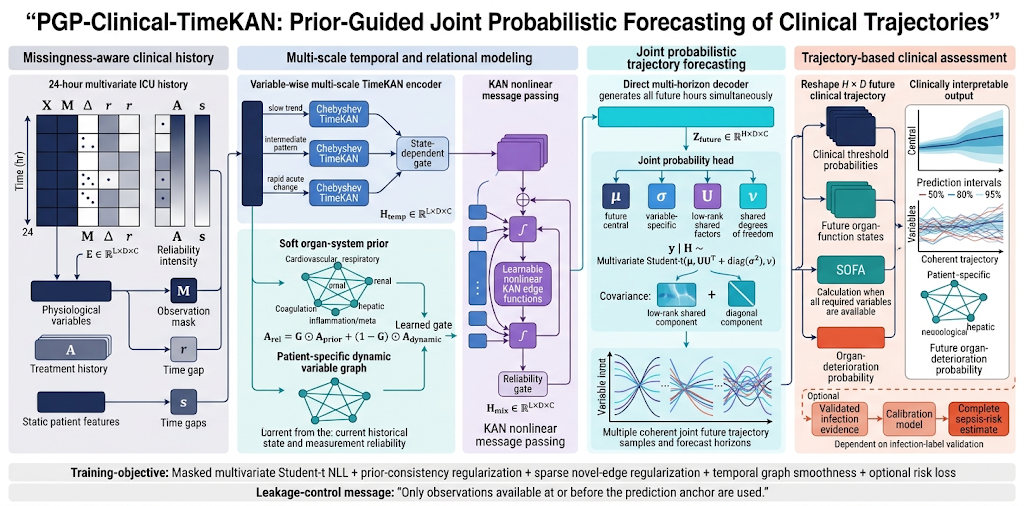}
    \caption{Overview of \method. Missingness-aware embeddings distinguish observed, imputed, and stale measurements. Variable-wise multi-scale TimeKAN encoders extract temporal patterns. A learned gate blends a soft organ-system prior with patient-specific relations, and KAN functions pass nonlinear messages across variables. A direct decoder parameterizes a low-rank multivariate Student-$t$ distribution over all variables and horizons. Coherent samples yield threshold-crossing and organ-deterioration probabilities. Complete sepsis-risk estimation requires independently validated infection evidence.}
    \label{fig:method-overview}
\end{figure}

Figure~\ref{fig:method-overview} shows the full \method{} pipeline. It maps leakage-controlled histories to joint trajectory samples and downstream organ-function estimates. The following sections define each component and the training objective.

\subsection{Missingness-aware clinical inputs}

For variable $d$ and historical hour $\tau$, the time since last observation is
\begin{equation}
\Delta_{\tau,d}=\begin{cases}
0,&M_{\tau,d}=1,\\
\Delta_{\tau-1,d}+1,&M_{\tau,d}=0.
\end{cases}
\end{equation}
We convert measurement age into a learned reliability
\begin{equation}
r_{\tau,d}=M_{\tau,d}+(1-M_{\tau,d})
\exp[-\operatorname{softplus}(\omega_d)\Delta_{\tau,d}],
\end{equation}
and embed $[\widetilde X_{\tau,d},M_{\tau,d},\log(1+\Delta_{\tau,d}),r_{\tau,d}]$ with variable identity, relative time, static covariates, and an optional treatment-history encoding. The representation distinguishes an imputed value from a contemporaneous measurement. It also keeps treatments separate from ordinary physiology.

\subsection{Variable-wise multi-scale TimeKAN encoder}

Each variable is encoded separately before cross-variable mixing. Let $\mathbf E_{:,d}\in\R^{L\times C}$ be its embedded history. A multi-scale decomposition produces $Q$ temporal components,
\begin{equation}
\mathbf E_{:,d}^{(q)}=\mathcal F^{-1}\!\left(\mathbf P_q\odot\mathcal F(\mathbf E_{:,d})\right),
\qquad q=1,\ldots,Q,
\end{equation}
where $\mathbf P_q$ selects a frequency range from the observed history. Each component is processed by a TimeKAN block. A scalar KAN edge uses a Chebyshev expansion
\begin{equation}
\phi(u)=\sum_{k=0}^{P}c_kT_k(u),
\quad T_0(u)=1,\quad T_1(u)=u,\quad
T_k(u)=2uT_{k-1}(u)-T_{k-2}(u).
\end{equation}
State-dependent gates combine the scale-specific representations:
\begin{equation}
\pi_{\tau,d}^{(q)}=\operatorname{softmax}_q g_q(\mathbf H_{\tau,d}^{(q)}),
\qquad
\mathbf H_{\tau,d}^{\mathrm{temp}}=\sum_q\pi_{\tau,d}^{(q)}\mathbf H_{\tau,d}^{(q)}.
\end{equation}
Short, nonstationary ICU trajectories may not favor Fourier decomposition. We therefore compare it with no decomposition, causal convolutions, and wavelets.

\subsection{Soft organ-system prior}

Let $\mathbf S\in\{0,1\}^{D\times K}$ map variables to $K$ organ or functional systems, allowing multiple memberships, and let $\mathbf C^{\mathrm{organ}}\in[0,1]^{K\times K}$ encode soft system-level relations. A variable-level prior is
\begin{align}
\widetilde{\mathbf A}^{\mathrm{prior}}
&=w_{\mathrm{self}}\mathbf I+w_{\mathrm{intra}}\mathbf S\mathbf S^\top
+w_{\mathrm{inter}}\mathbf S\mathbf C^{\mathrm{organ}}\mathbf S^\top,\\
\mathbf A^{\mathrm{prior}}&=\operatorname{RowNormalize}(\widetilde{\mathbf A}^{\mathrm{prior}}).
\end{align}
The graph is intentionally soft. It encourages clinically plausible routes without permanently excluding relations absent from the expert map.

\subsection{Patient-specific dynamic relations}

For each historical hour, query and key projections of $\mathbf H^{\mathrm{temp}}$ define
\begin{equation}
e_{\tau,dj}^{\mathrm{dyn}}=
\frac{\mathbf q_{\tau,d}^{\top}\mathbf k_{\tau,j}}{\sqrt C}
+b_{dj}^{\mathrm{rel}}+\log(r_{\tau,j}+\epsilon),
\quad
A_{\tau,dj}^{\mathrm{dyn}}=\operatorname{softmax}_j(e_{\tau,dj}^{\mathrm{dyn}}).
\end{equation}
A learned edge gate $G_{\tau,dj}\in(0,1)$ combines prior and data-driven relations:
\begin{equation}
\widetilde A_{\tau,dj}^{\mathrm{rel}}=
G_{\tau,dj}A_{dj}^{\mathrm{prior}}+(1-G_{\tau,dj})A_{\tau,dj}^{\mathrm{dyn}},
\quad
\mathbf A_\tau^{\mathrm{rel}}=\operatorname{RowNormalize}(\widetilde{\mathbf A}_\tau^{\mathrm{rel}}).
\end{equation}
An exponential update or smoothness penalty discourages implausible hourly changes in the graph. Every relation uses only the available history.

\subsection{KAN cross-variable message passing}

Attention weights express relation strength, but not the nonlinear form of a source variable's contribution. We therefore transform each source representation with a KAN edge function,
\begin{equation}
\vect\phi_{dj}^{\mathrm{KAN}}(\mathbf h)=
\sum_{p=0}^{P_r}\mathbf c_{dj,p}T_p(\tanh\mathbf h).
\end{equation}
To control parameters, the first implementation shares basis functions within organ-system pairs. Messages and residual updates are
\begin{align}
\mathbf m_{\tau,d}&=\sum_{j=1}^{D}\overline A_{\tau,dj}^{\mathrm{rel}}
\vect\phi_{dj}^{\mathrm{KAN}}(\mathbf H_{\tau,j}^{\mathrm{temp}}),\\
\mathbf H_{\tau,d}^{\mathrm{mix}}&=\operatorname{LayerNorm}\!\left(
\mathbf H_{\tau,d}^{\mathrm{temp}}+\vect\gamma_{\tau,d}\odot\mathbf m_{\tau,d}\right),
\end{align}
where $\vect\gamma_{\tau,d}$ depends on the target state, incoming message, and measurement reliability. One or two layers are used to limit over-smoothing.

\subsection{Direct multi-horizon decoder}

A direct decoder maps the mixed history and learned future queries to
$\mathbf Z_i^{\mathrm{future}}\in\R^{H\times D\times C}$. It predicts all horizons at once, avoiding autoregressive error accumulation. The joint probability head therefore receives the full $HD$-dimensional object.

\subsection{Low-rank multivariate Student-t output}

Let $\mathbf z_i=\operatorname{vec}(\mathbf Z_i^{\mathrm{future}})$ and $N=HD$. The output heads produce
\begin{align}
\vect\mu_i&=f_\mu(\mathbf z_i)\in\R^N, &
\vect\sigma_i&=\operatorname{softplus}(f_\sigma(\mathbf z_i))+\epsilon,\\
\mathbf U_i&=\operatorname{reshape}(f_U(\mathbf z_i))\in\R^{N\times R}, &
\nu_i&=2+\operatorname{softplus}(f_\nu(\mathbf z_i)).
\end{align}
With $R\ll N$, the joint scale matrix is
\begin{equation}
\vect\Sigma_i=\mathbf U_i\mathbf U_i^\top+\operatorname{diag}(\vect\sigma_i^2),
\qquad
\mathbf y_i\mid\mathcal H_{i,t}\sim t_{\nu_i}(\vect\mu_i,\vect\Sigma_i).
\end{equation}
The shared degree of freedom $\nu_i>2$ gives a well-defined multivariate Student-$t$ distribution with finite variance. The low-rank term captures deviations shared across variables and horizons. The diagonal term retains output-specific uncertainty.

The distribution admits a scale-mixture sampler. Drawing
$\tau_i\sim\operatorname{Gamma}(\nu_i/2,\nu_i/2)$,
$\mathbf z_i^s\sim\mathcal N(\mathbf0,\mathbf I_R/\tau_i)$, and
$\vect\epsilon_i\sim\mathcal N(\mathbf0,\mathbf I_N/\tau_i)$ yields
\begin{equation}
\mathbf y_i^{(s)}=\vect\mu_i+\mathbf U_i\mathbf z_i^s+
\vect\sigma_i\odot\vect\epsilon_i.
\end{equation}
Thus each sample is a coherent future course rather than a collection of independently drawn marginals.

\subsection{Masked joint likelihood and regularization}

Let $\mathcal O_i$ index the genuinely observed future targets and $n_i=|\mathcal O_i|$. Restricting the location and scale factors to those rows gives
\begin{equation}
\vect\Sigma_{i,\mathcal O}=\mathbf U_{i,\mathcal O}\mathbf U_{i,\mathcal O}^{\top}
+\operatorname{diag}(\vect\sigma_{i,\mathcal O}^2).
\end{equation}
The negative log-likelihood of the observed marginal is
\begin{align}
\mathcal L_{\mathrm{NLL}}^{(i)}={}&-\log\Gamma\!\left(\frac{\nu_i+n_i}{2}\right)
+\log\Gamma\!\left(\frac{\nu_i}{2}\right)+\frac{n_i}{2}\log(\nu_i\pi)
+\frac12\log|\vect\Sigma_{i,\mathcal O}|\\
&+\frac{\nu_i+n_i}{2}\log\!\left(1+
\frac{(\mathbf y_{i,\mathcal O}-\vect\mu_{i,\mathcal O})^\top
\vect\Sigma_{i,\mathcal O}^{-1}
(\mathbf y_{i,\mathcal O}-\vect\mu_{i,\mathcal O})}{\nu_i}\right).
\end{align}
The Woodbury identity and matrix determinant lemma move the main inverse and determinant computations into the $R$-dimensional factor space. The complete objective is
\begin{equation}
\mathcal L=\frac1B\sum_i\mathcal L_{\mathrm{NLL}}^{(i)}
+\lambda_p\mathcal L_{\mathrm{prior}}
+\lambda_s\mathcal L_{\mathrm{sparse}}
+\lambda_t\mathcal L_{\mathrm{smooth}}
+\lambda_r\mathcal L_{\mathrm{risk}},
\end{equation}
The first three regularizers discourage unsupported changes to confident prior edges, dense prior-external graphs, and abrupt temporal variation, respectively. Risk supervision is optional. We introduce it only after establishing trajectory calibration.

\subsection{From trajectory samples to clinical risk}

For a safe range $\mathcal R_d$, the probability of at least one threshold crossing is estimated by
\begin{equation}
\widehat p_{i,d}^{\mathrm{cross}}=\frac1S\sum_{s=1}^S
\mathbb I\!\left[\exists h:Y_{i,h,d}^{(s)}\notin\mathcal R_d\right].
\end{equation}
When all required respiratory, coagulation, liver, cardiovascular, neurological, and renal variables are available, each sampled future maps to SOFA~\citep{vincent1996sofa}. Otherwise, we report a prespecified observable-organ-function score and do not call it SOFA. Given baseline score $S_i^{\mathrm{base}}$, the deterioration probability is
\begin{equation}
\widehat p_i^{\mathrm{organ}}=\frac1S\sum_{s=1}^S
\mathbb I\!\left[\max_{1\le h\le H}\widehat S_{i,h}^{(s)}-S_i^{\mathrm{base}}\ge2\right].
\end{equation}
A validation-set calibrator combines infection evidence with $\widehat p_i^{\mathrm{organ}}$ only when suspected infection can be reconstructed reliably. It makes no independence assumption between infection and organ deterioration.

\section{Experiments}
\label{sec:experiments}

\subsection{Evaluation protocol}

The frozen experiment uses 24 hours of history to forecast six vital signs for the next six hours. The cohort contains 6,882 patients and 54,694 windows. The fixed patient-level split has 38,591 training, 8,116 validation, and 7,987 test windows. The test set includes 1,029 patients and 230,787 observed future targets. Table~\ref{tab:cohort} summarizes the cohort. The dataset-defined label is positive in 553 stays (8.0\% of patients). Positive windows are much rarer, with a prevalence of 1.039\% in the secondary risk task.

\begin{table}[t]
    \caption{Frozen cohort characteristics. Values are descriptive; no causal or group-difference claim is made.}
    \label{tab:cohort}
    \centering
    \small
    \begin{tabular}{lrrr}
        \toprule
        Characteristic & Overall & Positive & Negative \\
        \midrule
        Patients / stays & 6,882 & 553 & 6,329 \\
        Age, years & $64.39\pm15.45$ & $63.03\pm16.54$ & $64.51\pm15.35$ \\
        Female, \% & 40.74 & 47.56 & 40.15 \\
        Male, \% & 59.26 & 52.44 & 59.85 \\
        CVICU, \% & 34.19 & 13.38 & 36.01 \\
        \bottomrule
    \end{tabular}
\end{table}

All trainable models use the same split and observed-target masks. Model selection uses validation data only. We report the mean and standard deviation over five optimization seeds (20260823--20260827). Training uses AdamW, a learning rate of $10^{-3}$, weight decay of $10^{-4}$, and gradient-norm clipping at 1.0. We train for at most 80 epochs with batches of 256 and early-stopping patience of 10. For paired bootstrap comparisons, we resample the 1,029 test patients 2,000 times per seed. Resampling at the patient level preserves dependence among windows from the same patient.

We compare with persistence, Linear, DLinear, GRU-D, TCN, PatchTST, iTransformer, deterministic TimeKAN, and DeepVAR. We also include TimeKAN variants with independent Gaussian, independent Student-$t$, and low-rank Gaussian heads. Point metrics are masked normalized MAE and RMSE. Distributional metrics include marginal NLL, masked joint NLL per observed target, Monte Carlo CRPS, multivariate energy score, empirical coverage, and interval width. For secondary risk prediction, we compare with logistic regression, HistGradientBoosting, GRU-D, and iTransformer classifiers. All use the same histories and patient splits. We report AUROC, AUPRC, Brier score, ten-bin ECE, calibration slope and intercept, sensitivity at fixed alert rates, lead time, and decision-curve net benefit.

\subsection{Target observability}

\begin{figure}[t]
    \centering
    \includegraphics[width=\linewidth]{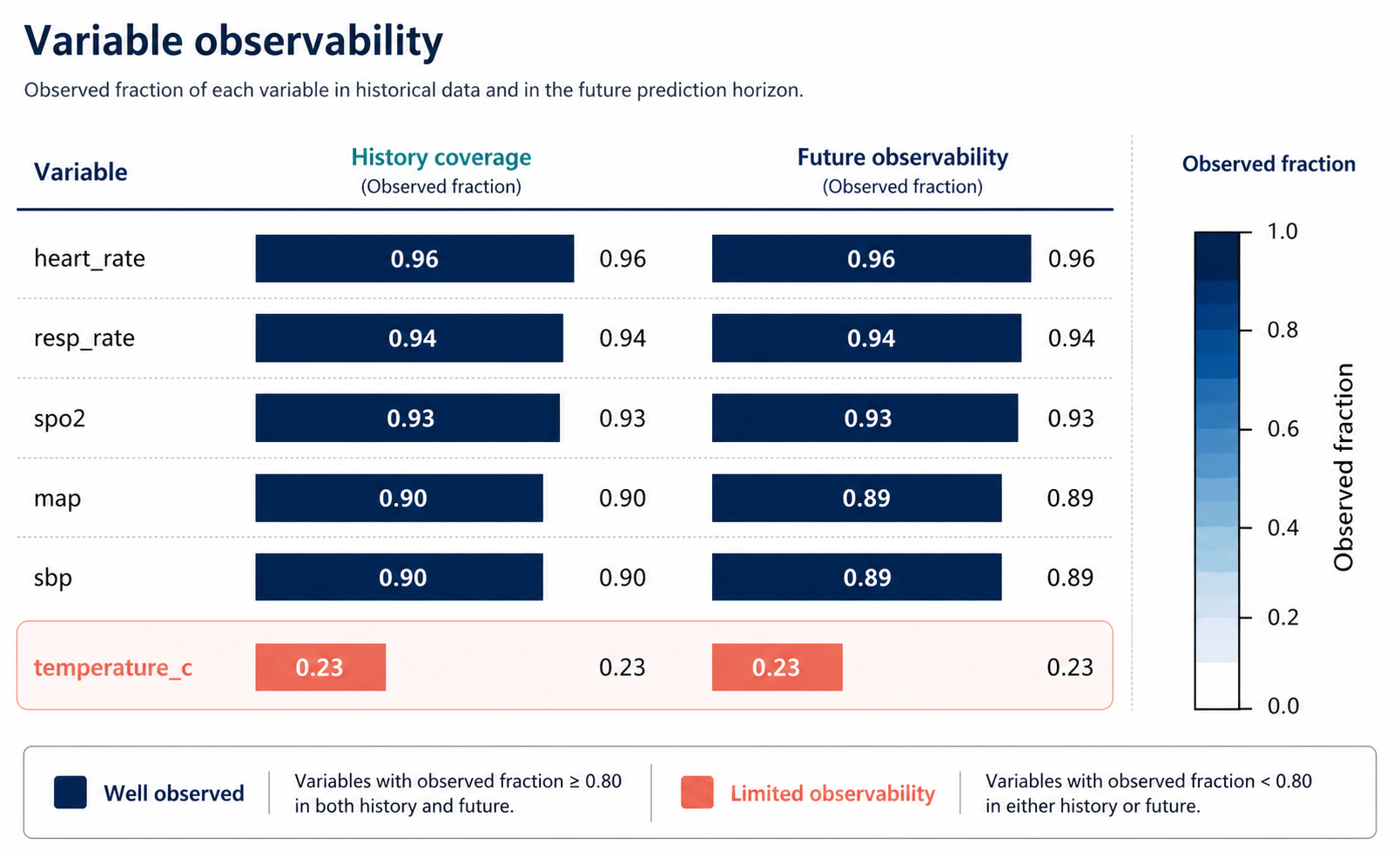}
    \caption{Variable-specific history coverage and future-target observability. Temperature is observed in about 23\% of bins, compared with more than 89\% for the other targets.}
    \label{fig:observability}
\end{figure}

Missingness varies sharply by variable (Figure~\ref{fig:observability}). Historical and future observability exceed 89\% for heart rate, respiratory rate, oxygen saturation, MAP, and SBP. For temperature, they are only 23.09\% and 23.20\%. We therefore evaluate every loss and metric on observed targets only. Temperature results have a smaller denominator and should not be compared with other variables by raw error alone.

\section{Results}
\label{sec:results}

\subsection{Point forecasting remains competitive}

Table~\ref{tab:point-main} reports the primary comparison. \method{} has the lowest RMSE, $0.52656\pm0.00034$, and the second-lowest MAE, $0.37727\pm0.00029$. GRU-D has a slightly lower MAE ($0.37636\pm0.00025$). The iTransformer result is close ($0.37750\pm0.00017$). Relative to deterministic TimeKAN, \method{} reduces MAE from 0.37923 to 0.37727, a 0.52\% relative change. RMSE falls from 0.53056 to 0.52656, a 0.75\% relative change. The relational and probabilistic components therefore preserve point accuracy. The results do not, however, support a state-of-the-art claim across all metrics.

\begin{table*}[t]
    \caption{Point forecasting on the frozen test set (five-seed mean $\pm$ standard deviation; lower is better). Bold denotes the best value and underline the second best.}
    \label{tab:point-main}
    \centering
    \small
    \resizebox{\textwidth}{!}{%
    \begin{tabular}{lcc@{\hspace{0.8cm}}lcc}
        \toprule
        Model & MAE $\downarrow$ & RMSE $\downarrow$ & Model & MAE $\downarrow$ & RMSE $\downarrow$ \\
        \midrule
        GRU-D & $\mathbf{0.37636}\pm0.00025$ & $0.52716\pm0.00082$ & DeepVAR & $0.38106\pm0.00035$ & $0.52752\pm0.00029$ \\
        \method{} & $\underline{0.37727}\pm0.00029$ & $\mathbf{0.52656}\pm0.00034$ & Independent Student-$t$ & $0.38107\pm0.00023$ & $0.53082\pm0.00010$ \\
        iTransformer & $0.37750\pm0.00017$ & $0.52789\pm0.00027$ & TCN & $0.38153\pm0.00030$ & $0.53232\pm0.00058$ \\
        TimeKAN & $0.37923\pm0.00028$ & $0.53056\pm0.00025$ & DLinear & $0.38162\pm0.00003$ & $0.53124\pm0.00007$ \\
        Linear & $0.38035\pm0.00008$ & $0.53025\pm0.00022$ & Independent Gaussian & $0.38260\pm0.00069$ & $0.53093\pm0.00066$ \\
        Low-rank Gaussian & $0.38306\pm0.00047$ & $0.53109\pm0.00033$ & PatchTST & $0.39684\pm0.00013$ & $0.54847\pm0.00038$ \\
        Persistence & $0.45190$ & $0.64734$ & & & \\
        \bottomrule
    \end{tabular}}
\end{table*}

\begin{figure}[t]
    \centering
    \includegraphics[width=\linewidth]{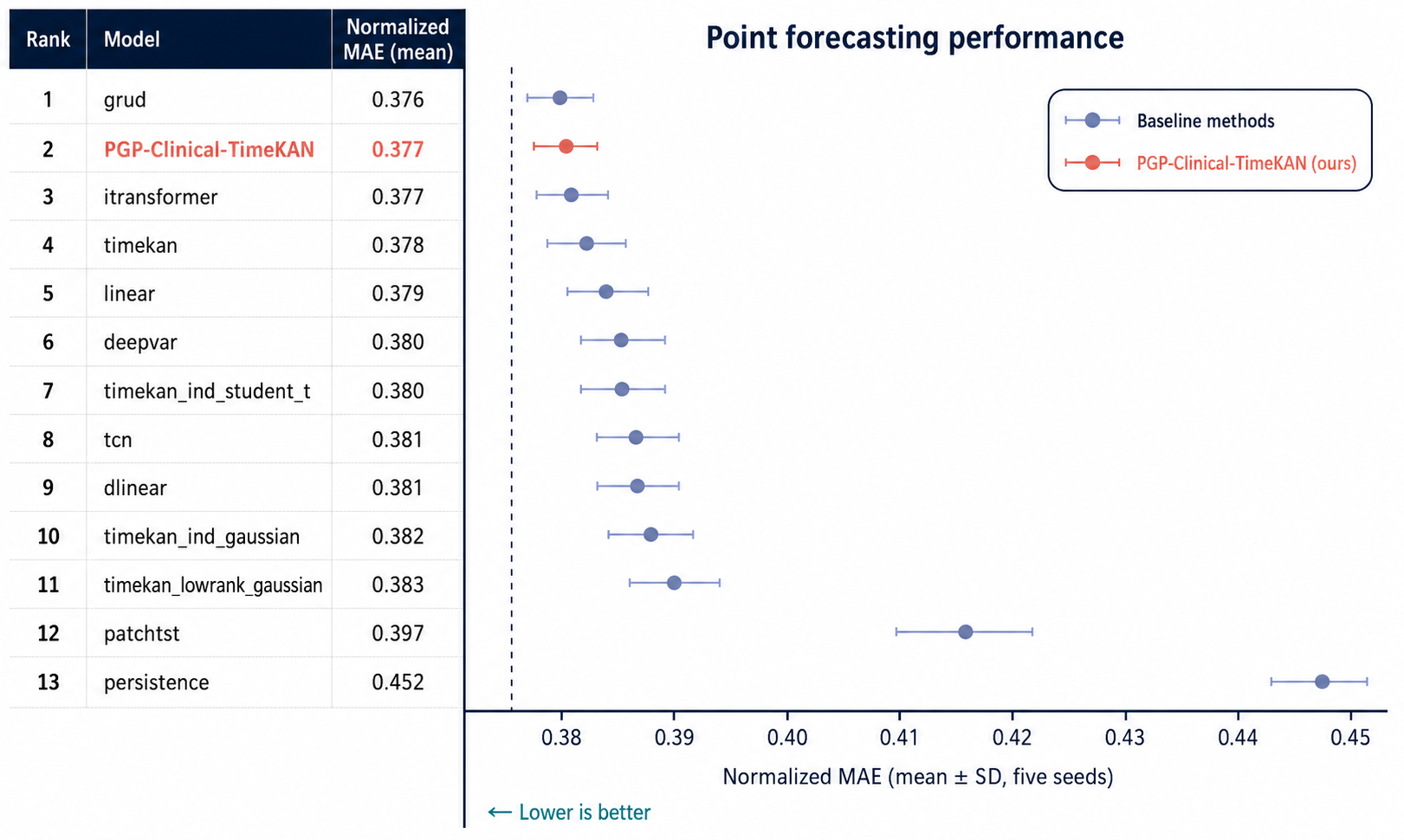}
    \caption{Normalized MAE across 13 forecasting models. \method{} belongs to the strongest group and ranks second on MAE; error bars show variation across five seeds.}
    \label{fig:point-performance}
\end{figure}

The paired bootstrap supports this bounded conclusion. We define each MAE difference as \method{} minus the comparator. Against TimeKAN, the mean difference is $-0.00196$, and every seed-specific 95\% interval is below zero. Differences also favor \method{} over Linear, DLinear, TCN, DeepVAR, PatchTST, and persistence. The result reverses against GRU-D: the difference is $+0.00091$, with every interval above zero. The interval against iTransformer crosses zero. Figure~\ref{fig:point-performance} shows the five-seed ranking.

\subsection{Error increases with horizon and acute variation}

\begin{figure*}[t]
    \centering
    \includegraphics[width=0.88\textwidth]{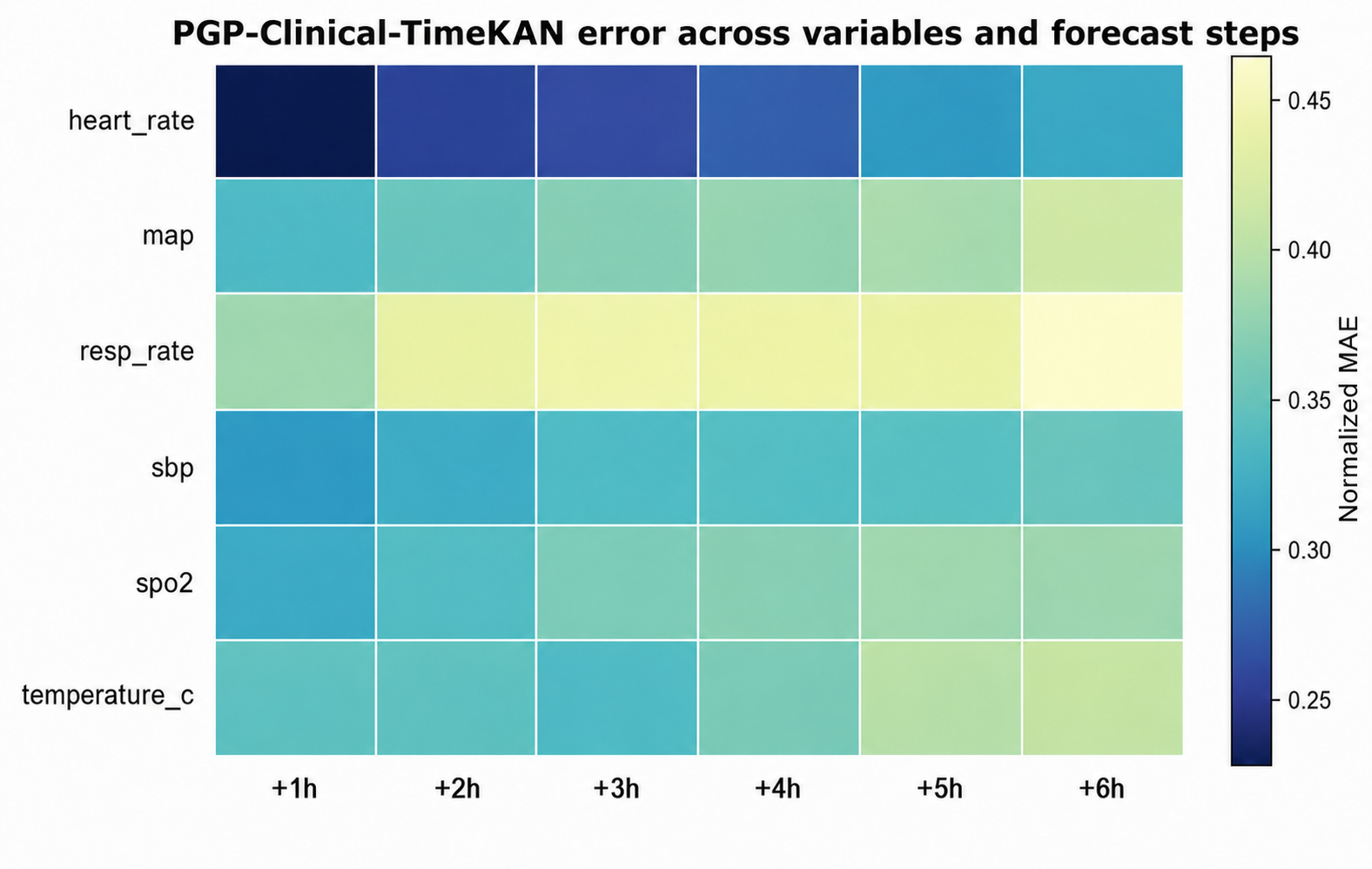}
    \caption{Normalized MAE by physiological variable and forecast step. Error generally increases with horizon, and a single aggregate score hides variation across variable--horizon pairs.}
    \label{fig:variable-horizon}
\end{figure*}

For \method{}, weighted MAE rises monotonically from 0.33784 at $+1$ hour to 0.40507 at $+6$ hours. RMSE increases from 0.48050 to 0.55455. The macro-average follows the same trend, so highly observed variables do not solely drive the degradation. In Figure~\ref{fig:variable-horizon}, the easiest variable--horizon pair is heart rate at $+1$ hour (MAE 0.22896). The hardest is respiratory rate at $+6$ hours (MAE 0.46076). Across variables, normalized MAE ranges from 0.28880 for heart rate to 0.44577 for respiratory rate. Appendix~\ref{app:detailed-results} reports errors in the original units.

\begin{figure*}[t]
    \centering
    \includegraphics[width=0.96\textwidth]{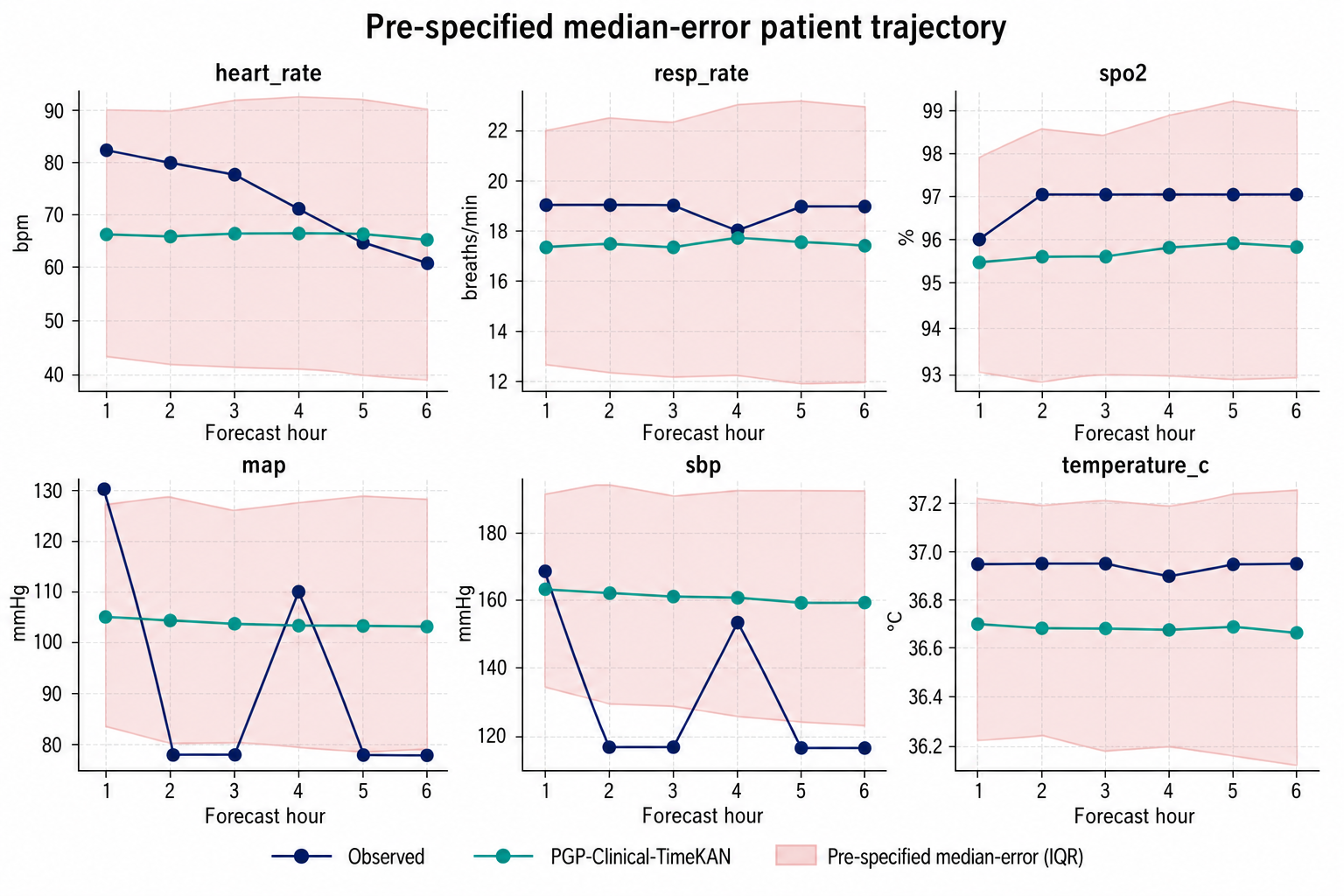}
    \caption{Prespecified median-error test trajectory. Lines show observations and forecast centers; shaded regions show the prespecified empirical error band. The uncertainty band is coherent, but the forecast smooths abrupt blood-pressure changes.}
    \label{fig:patient-trajectory}
\end{figure*}

The prespecified median-error case in Figure~\ref{fig:patient-trajectory} shows both utility and failure. Forecast centers are smooth and clinically plausible for stable trajectories. During abrupt MAP and SBP excursions, however, they regress toward the mean. Aggregate shape metrics show the same limitation. \method{} reaches a DTW of 0.3781 and a peak-time error of 1.792 hours, but is not uniformly better than GRU-D or Linear. Its direction accuracy is 0.4520, confirming that short-term turning points remain difficult.

\subsection{The joint Student-$t$ head improves distributional quality}

Table~\ref{tab:prob-main} compares the default rank-4 distribution with the main probabilistic baselines. \method{} has the lowest marginal NLL (0.66380), joint NLL per observation (0.56175), CRPS (0.27301), and energy score (0.35715). Empirical coverage is 0.533, 0.831, and 0.958 at the nominal 50\%, 80\%, and 95\% levels. The intervals are mildly conservative, especially at the two central levels; they are not perfectly calibrated. Independent Student-$t$ is closer to nominal at these levels but has worse proper scores. Coverage alone therefore does not capture forecast sharpness.

\begin{table*}[t]
    \caption{Probabilistic forecasting on 230,787 observed future targets. NLL, CRPS, and energy score are lower-is-better; coverage is reported for nominal 50/80/95\% intervals.}
    \label{tab:prob-main}
    \centering
    \small
    \resizebox{\textwidth}{!}{%
    \begin{tabular}{lccccc}
        \toprule
        Model & Marginal NLL $\downarrow$ & Joint NLL/obs $\downarrow$ & CRPS $\downarrow$ & Energy $\downarrow$ & Coverage 50/80/95 \\
        \midrule
        \method{} & \textbf{0.66380} & \textbf{0.56175} & \textbf{0.27301} & \textbf{0.35715} & 0.533 / 0.831 / 0.958 \\
        DeepVAR & 0.71282 & 0.61908 & 0.27559 & 0.35837 & 0.561 / 0.843 / 0.954 \\
        Independent Gaussian & 0.72296 & 0.72950 & 0.27683 & 0.36166 & 0.555 / 0.837 / 0.951 \\
        Independent Student-$t$ & 0.67036 & 0.67333 & 0.27547 & 0.36123 & 0.490 / 0.800 / 0.953 \\
        Low-rank Gaussian & 0.72618 & 0.66541 & 0.27783 & 0.36147 & 0.577 / 0.855 / 0.958 \\
        \bottomrule
    \end{tabular}}
\end{table*}

\begin{figure}[t]
    \centering
    \includegraphics[width=\linewidth]{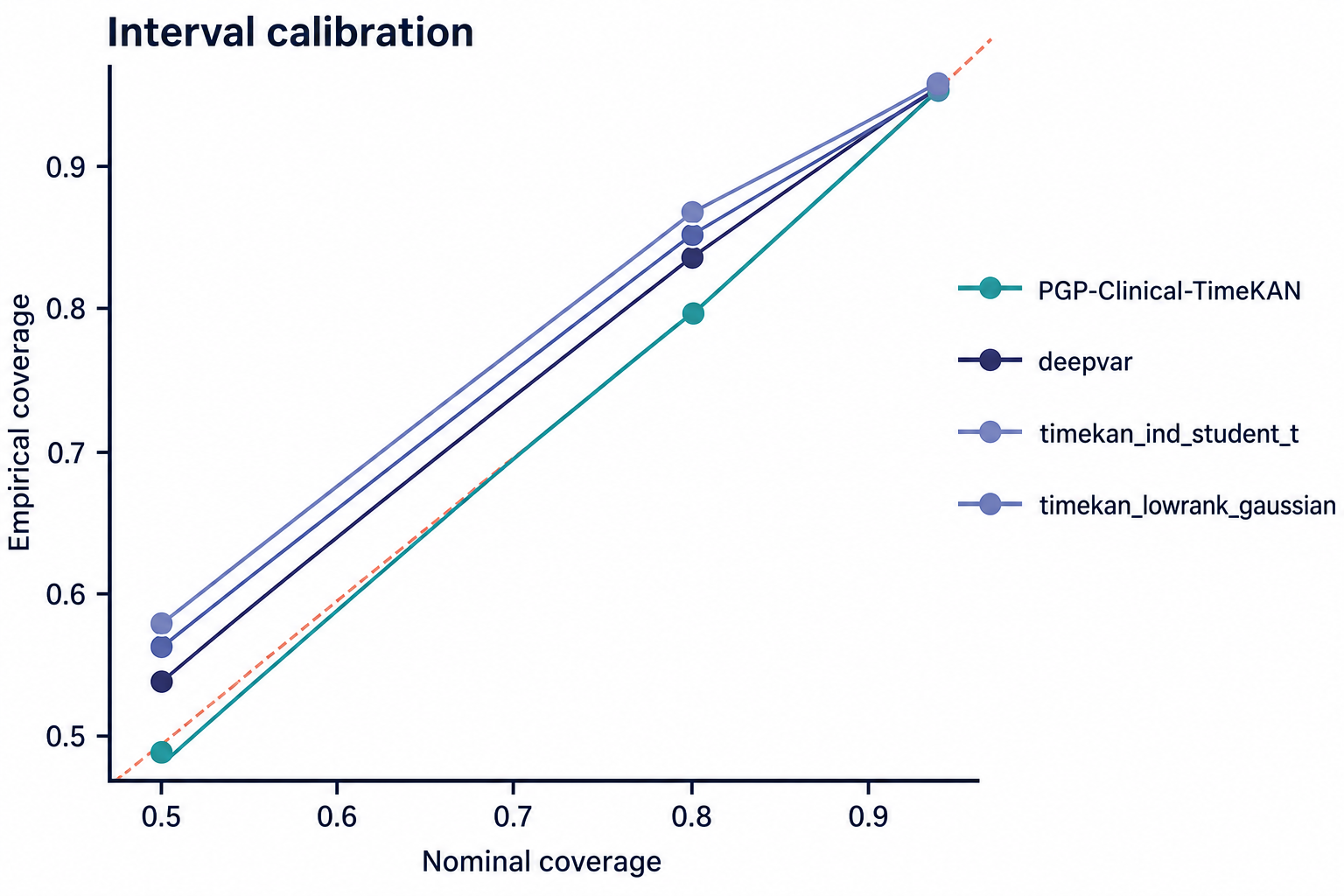}
    \caption{Nominal versus empirical interval coverage. \method{} is close to nominal at 95\% and mildly over-covers at 50\% and 80\%.}
    \label{fig:interval-calibration}
\end{figure}

\section{Ablation and further analysis}
\label{sec:ablation}

\subsection{Relational structure is the best-supported component}

\begin{figure*}[t]
    \centering
    \includegraphics[width=0.97\textwidth]{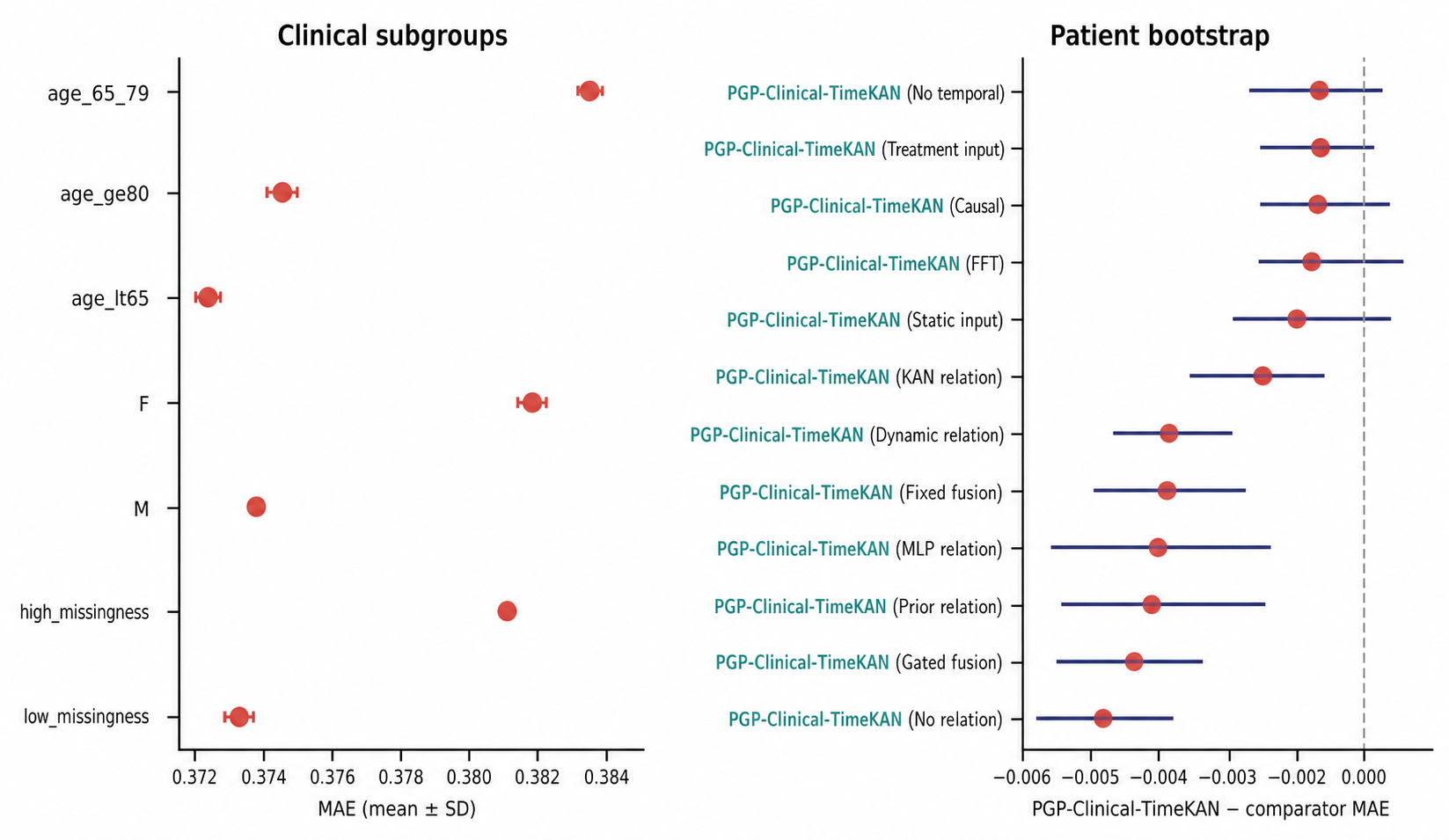}
    \caption{Clinical heterogeneity and patient-level ablations. Left: descriptive subgroup MAE estimates. Right: paired patient-bootstrap differences between the full model and ablations; negative values favor the full model. Small subgroups are exploratory and should not be interpreted as fairness conclusions.}
    \label{fig:subgroup-ablation}
\end{figure*}

Removing cross-variable relations causes the largest ablation loss (Table~\ref{tab:relation-main}). Among isolated variants, nonlinear KAN relations perform best and reduce MAE by 0.00293 relative to no relation. A prior-only graph is not sufficient. The full model also outperforms every isolated relation variant, so no single component explains the whole gain. Patient-level intervals in Figure~\ref{fig:subgroup-ablation} support the same result: relation ablations have larger effects than temporal-decomposition or missingness-input variants.

\begin{table}[t]
    \caption{Relation ablation (five-seed mean $\pm$ standard deviation).}
    \label{tab:relation-main}
    \centering
    \small
    \begin{tabular}{lcc}
        \toprule
        Relation setting & MAE $\downarrow$ & RMSE $\downarrow$ \\
        \midrule
        Full model & \textbf{$0.37727\pm0.00029$} & \textbf{0.52656} \\
        KAN relation only & $0.37869\pm0.00047$ & 0.52813 \\
        Dynamic relation & $0.38005\pm0.00022$ & 0.52969 \\
        Fixed fusion & $0.38011\pm0.00032$ & 0.52989 \\
        MLP relation & $0.38024\pm0.00065$ & 0.52865 \\
        Prior relation & $0.38035\pm0.00087$ & 0.52995 \\
        Gated fusion & $0.38077\pm0.00036$ & 0.53055 \\
        No relation & $0.38162\pm0.00045$ & 0.53051 \\
        \bottomrule
    \end{tabular}
\end{table}

\begin{figure*}[t]
    \centering
    \includegraphics[width=0.96\textwidth]{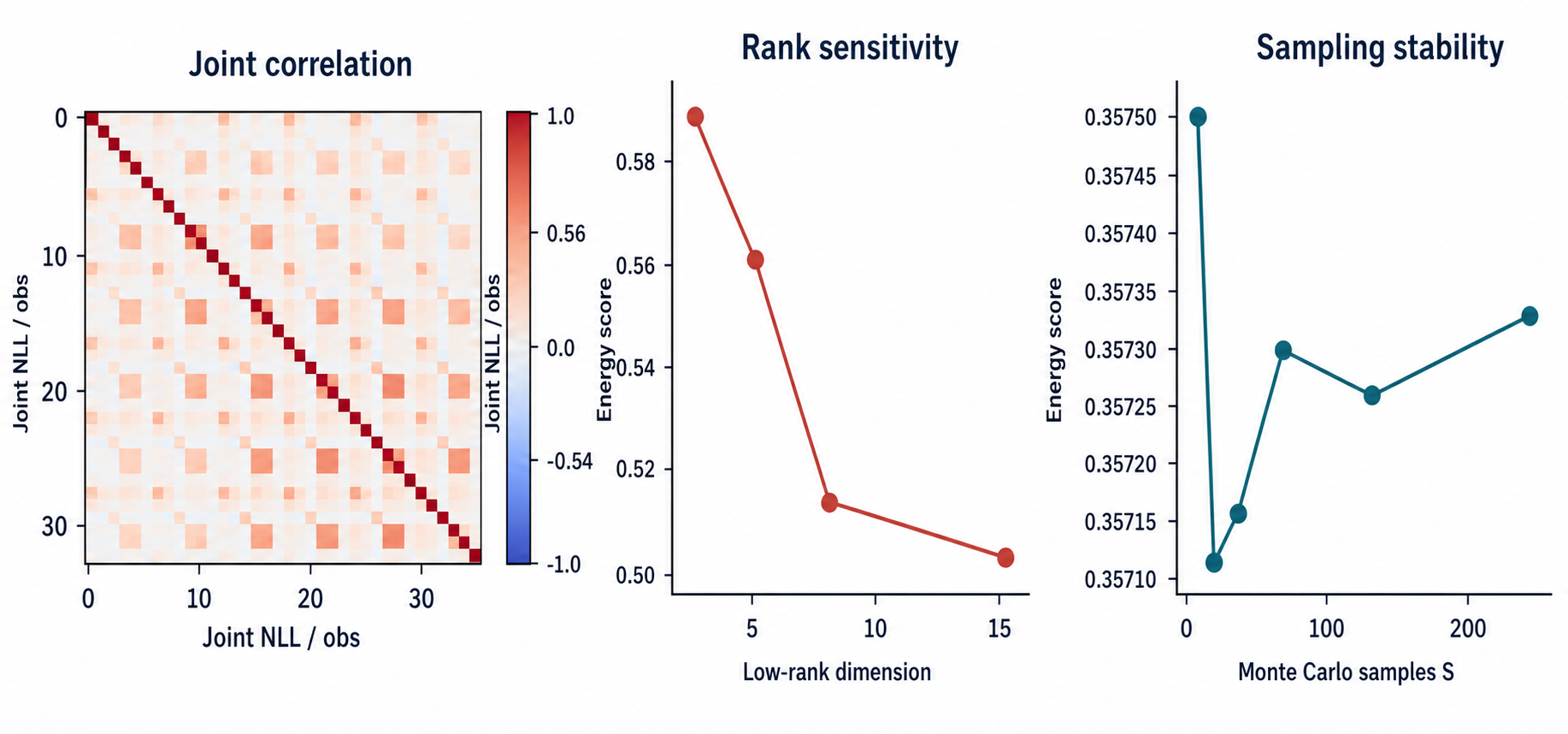}
    \caption{Joint dependence and numerical stability. Left: normalized predictive correlation across the 36 variable--horizon targets. Middle: higher covariance rank improves joint NLL with little change in MAE. Right: energy score is stable once the Monte Carlo sample count reaches 32.}
    \label{fig:joint-stability}
\end{figure*}

Raising the low-rank dimension from 2 to 16 changes MAE by at most 0.00026. In contrast, joint NLL per observed target improves from 0.58930 to 0.50323. Covariance rank therefore affects joint dependence more than the forecast center. Figure~\ref{fig:joint-stability} shows nonzero off-diagonal dependence, rank sensitivity, and stable Monte Carlo scores. Increasing the sample count from 8 to 256 moves CRPS only from 0.27351 into the 0.27301--0.27314 range. We use $S=32$ by default. The displayed matrix summarizes normalized predictive correlations; it is not a causal physiological graph.

\subsection{Sensitivity, subgroup heterogeneity, and efficiency}

Forecast length has the largest effect in the sensitivity study. MAE is 0.35860 at three hours, 0.37727 at six hours, 0.40155 at 12 hours, and 0.42832 at 24 hours. Extending the history from 24 to 48 hours does not help (MAE 0.38187). Values-only, values-plus-mask, and values-plus-mask-and-gap variants differ by less than $3\times10^{-5}$ MAE. Temporal variants range from 0.37749 to 0.37794. These small changes provide weaker evidence for the missingness and temporal branches than for cross-variable relations.

Errors are higher in several descriptive subgroups. High-missingness windows have an MAE of 0.38203 versus 0.37336 for low-missingness windows. Within 0--24 hours before a positive event, MAE is about 0.395, compared with 0.37639 for negative controls. Errors are also higher in TSICU and MICU/SICU. Some strata are very small, including 10 Neuro Stepdown and 26 Asian test patients. These estimates describe evaluation heterogeneity; they cannot establish demographic causation or fairness.

\method{} has about 0.51 million parameters. In the shared GPU microbenchmark, it takes 0.02757 ms per window and processes about 36,273 windows/s. Peak allocated memory is 84.94 MB. It is roughly 10 times slower than deterministic TimeKAN and 24 times slower than GRU-D. Even so, batched latency remains below 0.03 ms per window. The benchmark excludes data extraction, preprocessing, and clinical-system overhead.

\begin{figure*}[t]
    \centering
    \includegraphics[width=0.94\textwidth]{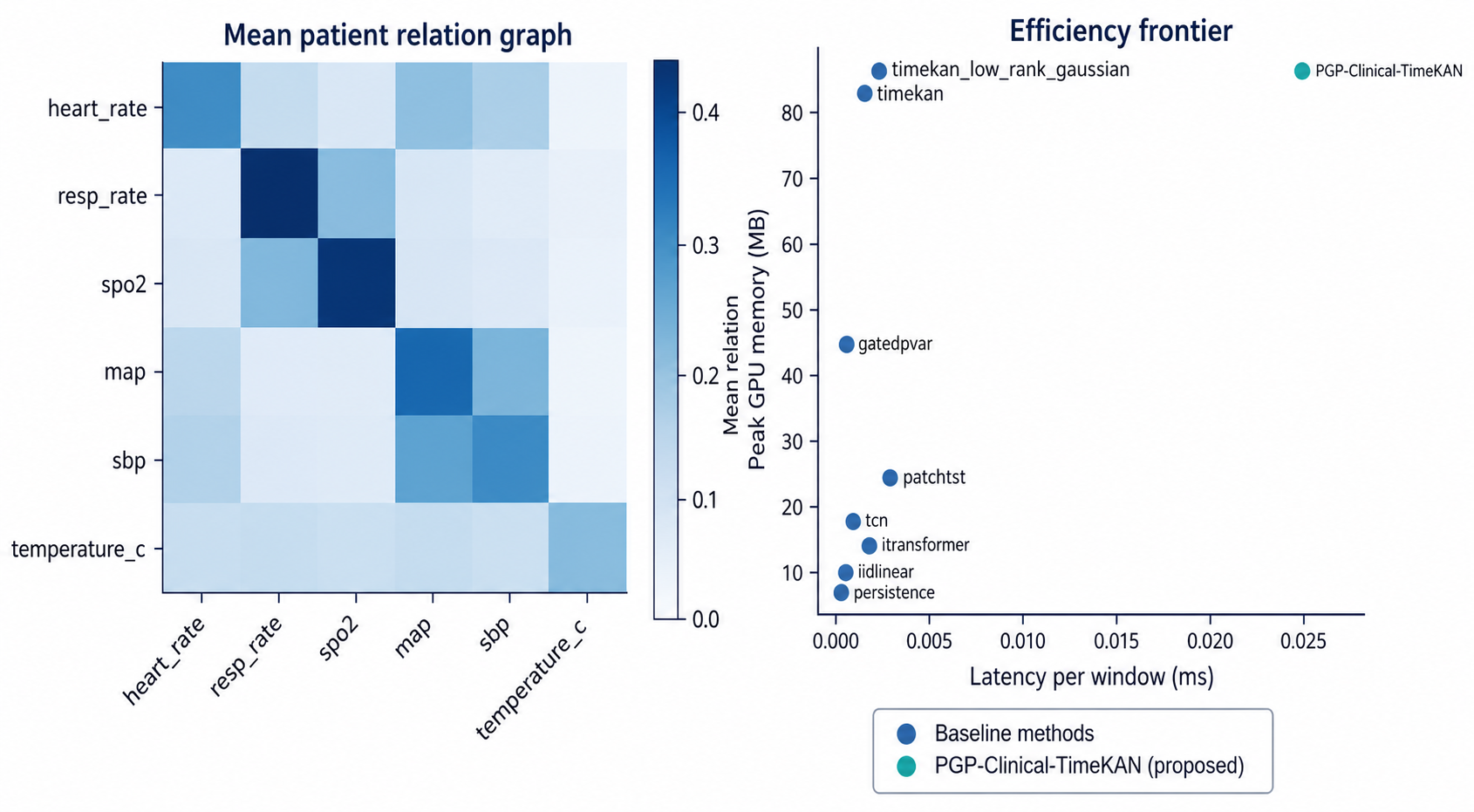}
    \caption{Mean predictive relation graph and efficiency frontier. Cross-variable associations, including MAP--SBP structure, are plausible predictive dependencies rather than causal mechanisms. Efficiency values are GPU microbenchmarks under a shared batch protocol, not end-to-end deployment latency.}
    \label{fig:graph-efficiency}
\end{figure*}

\subsection{Trajectory-derived risk is feasible but not competitive}

We evaluate trajectory-derived risk only as a secondary analysis (Table~\ref{tab:risk-main}). Its AUROC and AUPRC are 0.6030 and 0.0174, below the GRU-D classifier's 0.6503 and 0.0251. A calibration slope of 0.338 and ECE of 0.385 indicate compressed, poorly calibrated probabilities. At a 5\% window alert rate, sensitivity is 0.101. GRU-D reaches 0.157, and logistic regression reaches 0.205.

\begin{table*}[t]
    \caption{Secondary dataset-defined event prediction. Positive-window prevalence is 1.039\%; higher AUROC/AUPRC/sensitivity and lower Brier/ECE are better.}
    \label{tab:risk-main}
    \centering
    \small
    \resizebox{\textwidth}{!}{%
    \begin{tabular}{lccccc}
        \toprule
        Model & AUROC $\uparrow$ & AUPRC $\uparrow$ & Brier $\downarrow$ & ECE-10 $\downarrow$ & Sensitivity @ 5\% alerts $\uparrow$ \\
        \midrule
        GRU-D classifier & \textbf{0.6503} & \textbf{0.0251} & 0.1768 & \textbf{0.3653} & 0.1566 \\
        Logistic regression & 0.6402 & 0.0248 & 0.2211 & 0.4388 & \textbf{0.2048} \\
        iTransformer classifier & 0.6277 & 0.0199 & \textbf{0.1667} & 0.3680 & 0.1253 \\
        HistGradientBoosting & 0.6227 & 0.0195 & 0.1756 & 0.3961 & 0.0940 \\
        \method{} forecast--assess & 0.6030 & 0.0174 & 0.1996 & 0.3853 & 0.1012 \\
        \bottomrule
    \end{tabular}}
\end{table*}

\begin{figure*}[t]
    \centering
    \includegraphics[width=0.96\textwidth]{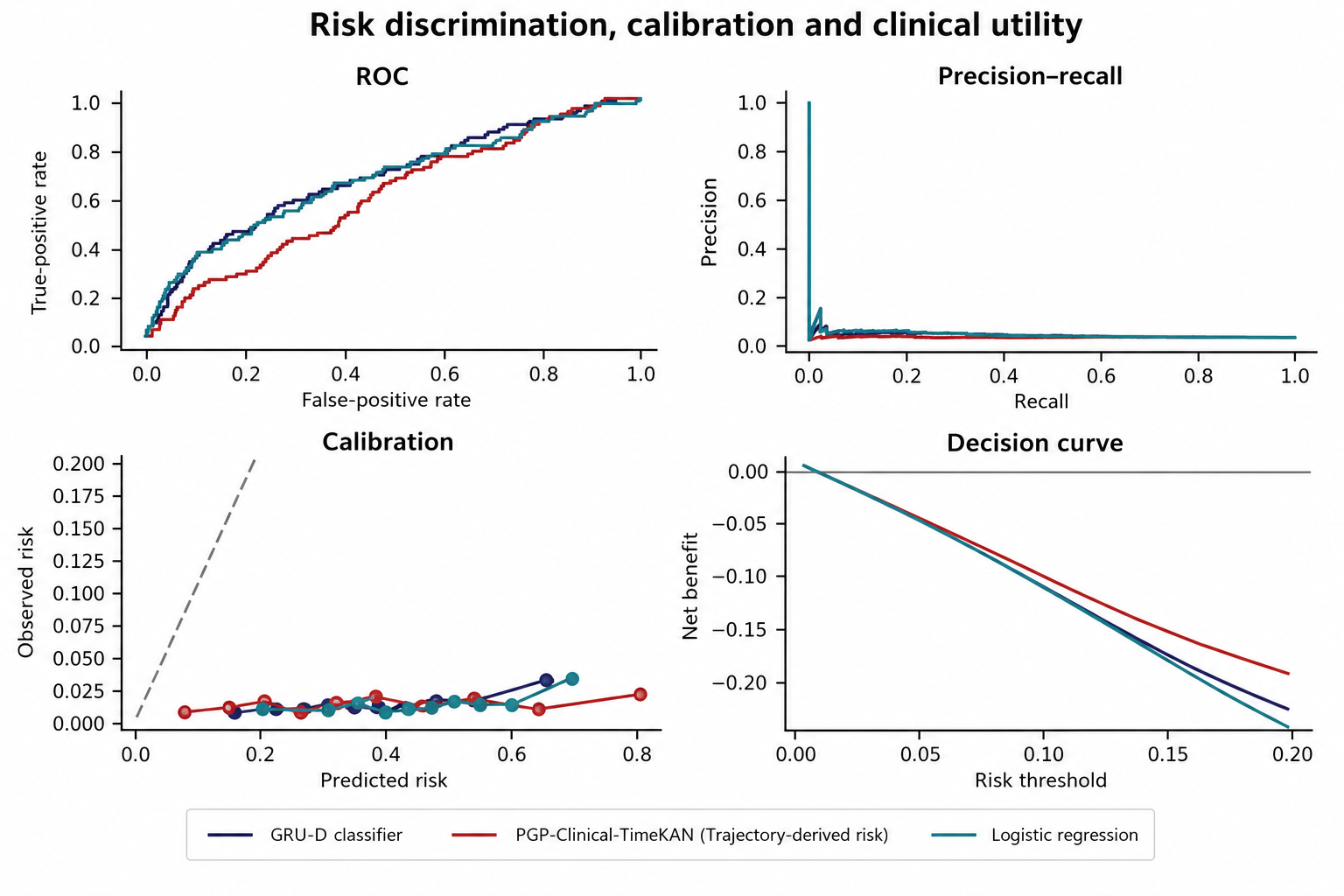}
    \caption{Discrimination, calibration, and decision curves for the secondary dataset-defined endpoint. The forecast--assess score is interpretable through trajectories but remains less discriminative and less well calibrated than dedicated classifiers.}
    \label{fig:risk-utility}
\end{figure*}

At event level, the trajectory-derived score detects 9.16\%, 32.29\%, and 49.16\% of positive stays at 1\%, 5\%, and 10\% alert rates. Median lead times are 13.9, 17.2, and 21.9 hours. Decision analysis finds a small positive net benefit at threshold 0.01 (0.000415), but negative benefit from 0.10 to 0.20. GRU-D is generally stronger. Matched negative pseudo-anchors raise AUPRC to 0.1505, but also change prevalence from 1.04\% to 8.07\%. That value is not comparable with the natural-window result. Figure~\ref{fig:risk-utility} documents feasibility and limitations, not clinical superiority.

\section{Discussion}

The results support a narrow trajectory-first claim. \method{} does not beat every point baseline. GRU-D retains a small, patient-level-consistent MAE advantage, and the paired bootstrap does not separate \method{} from iTransformer. Still, \method{} has the lowest RMSE and improves both MAE and RMSE over deterministic TimeKAN. Its joint heavy-tailed forecast also improves every reported proper score over the evaluated Gaussian, independent Student-$t$, and DeepVAR alternatives. The contribution is therefore not a universal point-forecasting gain. It is a competitive forecast center paired with a better representation of joint future uncertainty.

The ablations point to cross-variable relation modeling as the clearest source of improvement. Removing relations causes the largest MAE increase, and KAN relations outperform an MLP relation. Higher covariance rank improves joint likelihood with little change in point error. Together with nonzero off-diagonal predictive correlations, these results show that dependence across variables and horizons contributes to the forecast. They do not identify physiological causality. Learned edges mix biological coupling, measurement policy, treatment effects, and cohort-specific correlations. They should be read only as predictive associations.

Uncertainty quality is useful but imperfect. The 95\% interval is close to nominal coverage, whereas the 50\% and 80\% intervals are mildly conservative. Monte Carlo estimates stabilize at 32 samples, so sampling noise is unlikely to drive the main result. Two limitations remain visible. The representative case regresses toward the mean during abrupt blood-pressure changes, and error grows with forecast distance. A more flexible distribution might capture acute transitions better. Any gain, however, must justify the added training and sampling cost relative to the tractable observed-subset Student-$t$ likelihood.

The downstream risk experiment separates physiology forecasting from event classification. Coherent samples yield a traceable score with measurable lead time, but that score is less discriminative and less calibrated than dedicated classifiers. This negative result matters. Optimizing likelihood over future vital signs does not ensure that the distribution preserves rare features predictive of a terminal event. Future forecast--assess models may need endpoint-aware representations, richer organ variables, or joint training. They should still keep trajectory evidence separate from event labels.

Errors concentrate in clinically difficult settings. High-missingness windows, near-event periods, and some ICU strata have larger errors. The example trajectory also smooths acute hemodynamic changes. Global averages alone would hide these patterns. External validation should therefore cover different measurement policies and patient mixtures, with patient-level uncertainty for every subgroup estimate.

\section{Limitations, ethics, and broader impact}

This retrospective study uses a single-center database. The final task contains six vital signs, not the broader set of laboratory, treatment, and organ-function variables considered during method design. The supplied derivative also lacks a frozen, auditable time-varying treatment table. We therefore exclude treatment-input results from our claims. Static covariates are available, but their ablation gain is small.

The dataset-defined terminal-positive label is not an independently validated Sepsis-3 endpoint. The six-variable target set cannot reconstruct infection timing or the full six-organ SOFA state. We therefore avoid calling the secondary result a calibrated sepsis probability or complete SOFA deterioration. Positive-window prevalence is only 1.039\%, so AUPRC and alert metrics are sensitive to anchor construction. Matched pseudo-anchors alter prevalence and cannot show improvement under natural deployment conditions.

Subgroup analyses are exploratory. Some strata contain few patients, and observed differences may reflect sample size, disease mix, ICU workflow, or measurement frequency rather than demographic effects. These results do not establish fairness or causal group disparities. External, temporal, and multi-site validation is required before clinical use.

False reassurance and alert fatigue are plausible harms. The trajectory-derived score is not accurate or calibrated enough for deployment. The GPU benchmark also omits extraction, preprocessing, interface, and clinician-response delays. A prospective system should display uncertainty and support abstention or escalation for out-of-distribution inputs. It should retain human oversight and report patient-level outcomes and alerts per patient-hour. We use deidentified MIMIC-IV-derived data under its access conditions and redistribute no identifiable records.

\section{Conclusion}

We presented \method, a prior-guided framework for joint probabilistic forecasting of short-horizon clinical trajectories. On a frozen cohort of 6,882 patients, it obtains the lowest RMSE and second-lowest MAE among 13 models. It also improves consistently over deterministic TimeKAN and achieves the best evaluated marginal NLL, joint NLL, CRPS, and energy score. Relation ablations and covariance-rank analyses give the strongest support to cross-variable dependence. The resulting trajectories form an inspectable intermediate representation. However, the secondary event score remains weaker than dedicated classifiers and is not a clinically validated sepsis predictor. Joint trajectory modeling is therefore useful in its own right, but clinical translation still requires stronger endpoint evidence, calibration, and data coverage.

\bibliographystyle{plainnat}
\bibliography{references}

\appendix
\clearpage
\section{Detailed experimental results}
\label{app:detailed-results}

\setcounter{table}{0}
\renewcommand{\thetable}{A\arabic{table}}
\renewcommand{\theHtable}{A\arabic{table}}

\begin{table}[h]
    \caption{Variable observability in the frozen test set.}
    \label{tab:observability-detail}
    \centering
    \small
    \begin{tabular}{lrrrr}
        \toprule
        Variable & Unit & History (\%) & Future (\%) & Future $n$ \\
        \midrule
        Heart rate & beats/min & 95.55 & 95.92 & 45,912 \\
        Respiratory rate & breaths/min & 94.18 & 94.39 & 45,048 \\
        SpO$_2$ & \% & 93.08 & 92.59 & 44,317 \\
        MAP & mmHg & 90.13 & 89.44 & 42,271 \\
        SBP & mmHg & 90.15 & 89.47 & 42,286 \\
        Temperature & $^\circ$C & 23.09 & 23.20 & 10,953 \\
        \bottomrule
    \end{tabular}
\end{table}

\begin{table}[h]
    \caption{Variable-level \method{} performance. Original-unit errors are not comparable across variables.}
    \label{tab:variable-detail}
    \centering
    \small
    \resizebox{\columnwidth}{!}{%
    \begin{tabular}{lrrrr}
        \toprule
        Variable & Observed $n$ & Norm. MAE & Norm. RMSE & Original-unit MAE \\
        \midrule
        Heart rate & 45,912 & 0.28880 & 0.41376 & 6.64 beats/min \\
        MAP & 42,271 & 0.40369 & 0.55492 & 7.67 mmHg \\
        Respiratory rate & 45,048 & 0.44577 & 0.60735 & 3.12 breaths/min \\
        SBP & 42,286 & 0.35739 & 0.47164 & 10.72 mmHg \\
        SpO$_2$ & 44,317 & 0.38762 & 0.55539 & 1.55 percentage points \\
        Temperature & 10,953 & 0.39928 & 0.55876 & 0.266 $^\circ$C \\
        \bottomrule
    \end{tabular}}
\end{table}

\begin{table}[h]
    \caption{Horizon-level \method{} performance.}
    \label{tab:horizon-detail}
    \centering
    \small
    \begin{tabular}{rrrrr}
        \toprule
        Horizon & Observed $n$ & Weighted MAE & Weighted RMSE & Macro MAE \\
        \midrule
        $+1$ h & 38,637 & 0.33784 & 0.48050 & 0.34414 \\
        $+2$ h & 38,464 & 0.36070 & 0.51120 & 0.36419 \\
        $+3$ h & 38,708 & 0.37692 & 0.52820 & 0.37675 \\
        $+4$ h & 38,344 & 0.38674 & 0.53442 & 0.38798 \\
        $+5$ h & 38,476 & 0.39676 & 0.54747 & 0.40159 \\
        $+6$ h & 38,158 & 0.40507 & 0.55455 & 0.40851 \\
        \bottomrule
    \end{tabular}
\end{table}

\begin{table*}[t]
    \caption{Trajectory-shape metrics (five-seed mean $\pm$ standard deviation). Direction accuracy is higher-is-better; peak-time error and DTW are lower-is-better.}
    \label{tab:shape-detail}
    \centering
    \small
    \begin{tabular}{lccc}
        \toprule
        Model & Direction accuracy & Peak-time error (h) & DTW \\
        \midrule
        DeepVAR & $0.4489\pm0.0011$ & $1.730\pm0.033$ & $0.3809\pm0.0003$ \\
        DLinear & $\mathbf{0.4533}\pm0.0007$ & $1.846\pm0.029$ & $0.3829\pm0.0001$ \\
        GRU-D & $0.4502\pm0.0006$ & $1.783\pm0.019$ & $\mathbf{0.3764}\pm0.0003$ \\
        iTransformer & $0.4506\pm0.0004$ & $1.746\pm0.059$ & $0.3774\pm0.0002$ \\
        Linear & $0.4492\pm0.0011$ & $\mathbf{1.724}\pm0.008$ & $0.3784\pm0.0001$ \\
        PatchTST & $0.4452\pm0.0017$ & $1.840\pm0.168$ & $0.3973\pm0.0004$ \\
        Persistence & 0.1179 & 1.960 & 0.4553 \\
        \method{} & $0.4520\pm0.0010$ & $1.792\pm0.075$ & $0.3781\pm0.0003$ \\
        TCN & $0.4488\pm0.0008$ & $1.738\pm0.026$ & $0.3815\pm0.0003$ \\
        TimeKAN & $0.4520\pm0.0020$ & $1.856\pm0.063$ & $0.3802\pm0.0003$ \\
        Independent Gaussian & $0.4519\pm0.0005$ & $1.847\pm0.049$ & $0.3839\pm0.0007$ \\
        Independent Student-$t$ & $0.4513\pm0.0011$ & $1.831\pm0.076$ & $0.3824\pm0.0003$ \\
        Low-rank Gaussian & $0.4509\pm0.0012$ & $1.773\pm0.038$ & $0.3839\pm0.0004$ \\
        \bottomrule
    \end{tabular}
\end{table*}

\begin{table*}[t]
    \caption{Paired patient-level bootstrap. $\Delta$ is \method{} MAE minus comparator MAE, so negative values favor \method{}. Bounds are the lowest and highest seed-specific 95\% interval endpoints.}
    \label{tab:bootstrap-detail}
    \centering
    \small
    \resizebox{\textwidth}{!}{%
    \begin{tabular}{lrrrl}
        \toprule
        Comparator & Mean $\Delta$ & Lowest bound & Highest bound & Interpretation \\
        \midrule
        DeepVAR & -0.003793 & -0.004790 & -0.002632 & \method{} lower in all seeds \\
        DLinear & -0.004345 & -0.005359 & -0.003373 & \method{} lower in all seeds \\
        GRU-D & +0.000908 & +0.000100 & +0.002222 & GRU-D lower in all seeds \\
        iTransformer & -0.000232 & -0.001204 & +0.000736 & At least one interval crosses zero \\
        Linear & -0.003081 & -0.004106 & -0.002097 & \method{} lower in all seeds \\
        PatchTST & -0.019574 & -0.021542 & -0.017902 & \method{} lower in all seeds \\
        Persistence & -0.074627 & -0.078193 & -0.071228 & \method{} lower in all seeds \\
        TCN & -0.004263 & -0.005729 & -0.002955 & \method{} lower in all seeds \\
        TimeKAN & -0.001963 & -0.002838 & -0.000902 & \method{} lower in all seeds \\
        \bottomrule
    \end{tabular}}
\end{table*}

\begin{table}[h]
    \caption{Low-rank covariance sensitivity.}
    \label{tab:rank-detail}
    \centering
    \small
    \begin{tabular}{rrrr}
        \toprule
        Rank & MAE & RMSE & Joint NLL/obs \\
        \midrule
        2 & 0.37756 & 0.52685 & 0.58930 \\
        4 (default) & \textbf{0.37727} & 0.52656 & 0.56175 \\
        8 & 0.37771 & 0.52649 & 0.52746 \\
        16 & 0.37745 & \textbf{0.52650} & \textbf{0.50323} \\
        \bottomrule
    \end{tabular}
\end{table}

\begin{table*}[t]
    \caption{Input and temporal ablations. Differences in MAE are small relative to relation ablations.}
    \label{tab:input-temporal-detail}
    \centering
    \small
    \begin{tabular}{llccc}
        \toprule
        Suite & Setting & MAE & RMSE & Parameters \\
        \midrule
        Input & Values only & 0.377663 & 0.526623 & 507,395 \\
        Input & Values + mask & 0.377681 & 0.526710 & 507,395 \\
        Input & Values + mask + gap & 0.377685 & 0.526659 & 507,395 \\
        Input & Static covariates & 0.378181 & 0.526588 & 513,923 \\
        \midrule
        Temporal & Wavelet & \textbf{0.377490} & 0.526109 & 506,627 \\
        Temporal & None & 0.377779 & 0.526457 & 506,241 \\
        Temporal & Causal convolution & 0.377837 & 0.526893 & 507,395 \\
        Temporal & FFT & 0.377936 & \textbf{0.525859} & 506,627 \\
        \bottomrule
    \end{tabular}
\end{table*}

\begin{table}[h]
    \caption{History, horizon, and anchor-stride sensitivity. Settings use notation history--prediction--stride in hours.}
    \label{tab:sensitivity-detail}
    \centering
    \small
    \begin{tabular}{lcc}
        \toprule
        Setting & MAE & RMSE \\
        \midrule
        h24--p3--s6 & 0.35860 & 0.50842 \\
        h24--p6--s6 & 0.37727 & 0.52656 \\
        h24--p12--s6 & 0.40155 & 0.55310 \\
        h24--p24--s6 & 0.42832 & 0.58229 \\
        h48--p6--s6 & 0.38187 & 0.53512 \\
        h12--p6--s6 & 0.37705 & 0.52630 \\
        h24--p6--s3 & 0.37791 & 0.52728 \\
        h24--p6--s12 & 0.38041 & 0.52877 \\
        \bottomrule
    \end{tabular}
\end{table}

\begin{table*}[t]
    \caption{Clinical-range sensitivity. The strict profile removes only a small number of targets; the wide profile is identical to the protocol profile.}
    \label{tab:range-detail}
    \centering
    \small
    \begin{tabular}{lrrrrr}
        \toprule
        Variable & Protocol $n$ & Strict $n$ & Protocol MAE & Strict MAE & Protocol / strict RMSE \\
        \midrule
        Heart rate & 45,912 & 45,912 & 0.28880 & 0.28880 & 0.41376 / 0.41376 \\
        MAP & 42,271 & 42,245 & 0.40369 & 0.40174 & 0.55492 / 0.54697 \\
        Respiratory rate & 45,048 & 45,036 & 0.44577 & 0.44468 & 0.60735 / 0.60157 \\
        SBP & 42,286 & 42,276 & 0.35739 & 0.35706 & 0.47164 / 0.47090 \\
        SpO$_2$ & 44,317 & 44,304 & 0.38762 & 0.38531 & 0.55539 / 0.53688 \\
        Temperature & 10,953 & 10,953 & 0.39928 & 0.39928 & 0.55876 / 0.55876 \\
        \bottomrule
    \end{tabular}
\end{table*}

\begin{table*}[t]
    \caption{Descriptive subgroup analysis. Small strata are retained for transparency but are not used for fairness or causal claims.}
    \label{tab:subgroup-detail}
    \centering
    \scriptsize
    \begin{tabular}{llrrrcc}
        \toprule
        Type & Subgroup & Patients & Windows & Seeds & MAE & RMSE \\
        \midrule
        Age & $<65$ & 466 & 3,725 & 5 & 0.37241 & 0.51711 \\
        Age & 65--79 & 390 & 2,946 & 5 & 0.38434 & 0.53710 \\
        Age & $\geq80$ & 173 & 1,316 & 5 & 0.37485 & 0.52861 \\
        Gender & Female & 394 & 3,041 & 5 & 0.38272 & 0.53775 \\
        Gender & Male & 635 & 4,946 & 5 & 0.37394 & 0.51959 \\
        Missingness & High & 807 & 3,839 & 5 & 0.38203 & 0.53176 \\
        Missingness & Low & 728 & 4,148 & 5 & 0.37336 & 0.52226 \\
        Disease phase & Negative control & 946 & 7,363 & 5 & 0.37639 & 0.52500 \\
        Disease phase & Positive 0--6 h & 83 & 164 & 5 & 0.39546 & 0.55334 \\
        Disease phase & Positive 7--24 h & 65 & 164 & 5 & 0.39515 & 0.54478 \\
        Disease phase & Positive $>24$ h & 40 & 296 & 5 & 0.37850 & 0.53842 \\
        ICU & CCU & 162 & 1,235 & 5 & 0.36335 & 0.50148 \\
        ICU & CVICU & 348 & 2,241 & 5 & 0.37020 & 0.51468 \\
        ICU & MICU & 91 & 504 & 5 & 0.38218 & 0.53236 \\
        ICU & MICU/SICU & 113 & 654 & 5 & 0.39876 & 0.56131 \\
        ICU & Neuro Intermediate & 29 & 285 & 5 & 0.36544 & 0.49613 \\
        ICU & Neuro SICU & 43 & 694 & 5 & 0.38667 & 0.56795 \\
        ICU & Neuro Stepdown & 10 & 109 & 5 & 0.34064 & 0.49289 \\
        ICU & SICU & 105 & 1,181 & 5 & 0.36521 & 0.50433 \\
        ICU & TSICU & 128 & 1,084 & 5 & 0.40493 & 0.55963 \\
        Race & Asian & 26 & 265 & 5 & 0.35083 & 0.48141 \\
        Race & Black & 95 & 877 & 5 & 0.37128 & 0.51867 \\
        Race & Other/missing & 164 & 1,318 & 5 & 0.37063 & 0.52001 \\
        Race & White & 744 & 5,527 & 5 & 0.38107 & 0.53141 \\
        \bottomrule
    \end{tabular}
\end{table*}

\begin{table}[h]
    \caption{Monte Carlo sampling stability. Coverage is invariant because it is computed analytically from the fitted distribution.}
    \label{tab:sampling-detail}
    \centering
    \small
    \begin{tabular}{rccc}
        \toprule
        Samples $S$ & Joint NLL/obs & CRPS & Energy \\
        \midrule
        8 & 0.56175 & 0.27351 & 0.35749 \\
        16 & 0.56175 & 0.27304 & 0.35711 \\
        32 & 0.56175 & \textbf{0.27301} & 0.35715 \\
        64 & 0.56175 & 0.27312 & 0.35729 \\
        128 & 0.56175 & 0.27310 & 0.35725 \\
        256 & 0.56175 & 0.27314 & 0.35732 \\
        \bottomrule
    \end{tabular}
\end{table}

\begin{table*}[t]
    \caption{Efficiency microbenchmark under a common GPU batch protocol. Values exclude data loading and clinical-system overhead.}
    \label{tab:efficiency-detail}
    \centering
    \small
    \begin{tabular}{lrrrr}
        \toprule
        Model & Parameters & ms/window & Windows/s & Peak allocated MB \\
        \midrule
        Persistence & 1 & 0.000126 & 7,949,337 & 8.58 \\
        Linear & 5,220 & 0.000123 & 8,146,056 & 9.60 \\
        DLinear & 300 & 0.000507 & 1,972,675 & 9.76 \\
        GRU-D & 18,510 & 0.001134 & 882,164 & 44.10 \\
        TCN & 30,564 & 0.001211 & 825,707 & 17.63 \\
        DeepVAR & 43,428 & 0.001387 & 721,161 & 43.92 \\
        iTransformer & 105,030 & 0.002156 & 463,878 & 13.75 \\
        TimeKAN & 8,329 & 0.002687 & 372,120 & 83.02 \\
        PatchTST & 101,574 & 0.003511 & 284,779 & 24.28 \\
        \method{} & 507,395 & 0.027569 & 36,273 & 84.94 \\
        \bottomrule
    \end{tabular}
\end{table*}

\begin{table*}[t]
    \caption{Extended secondary-risk results. Calibration is measured at the window level. Sensitivity and lead time are episode-level quantities and use a different denominator.}
    \label{tab:risk-extended-detail}
    \centering
    \small
    \resizebox{\textwidth}{!}{%
    \begin{tabular}{lrrrrrr}
        \toprule
        Model & AUROC & AUPRC & Calibration intercept & Calibration slope & Brier & ECE-10 \\
        \midrule
        GRU-D classifier & 0.6503 & 0.0251 & -4.331 & 0.656 & 0.1768 & 0.3653 \\
        Logistic regression & 0.6402 & 0.0248 & -4.491 & 0.719 & 0.2211 & 0.4388 \\
        iTransformer classifier & 0.6277 & 0.0199 & -4.251 & 0.907 & 0.1667 & 0.3680 \\
        HistGradientBoosting & 0.6227 & 0.0195 & -4.229 & 1.071 & 0.1756 & 0.3961 \\
        \method{} forecast--assess & 0.6030 & 0.0174 & -4.449 & 0.338 & 0.1996 & 0.3853 \\
        \bottomrule
    \end{tabular}}

    \vspace{0.6em}
    \begin{tabular}{rrrr}
        \toprule
        Alert rate & Episode sensitivity & Median lead time (h) & Interpretation \\
        \midrule
        1\% & 0.0916 & 13.9 & Low alert burden, few events detected \\
        5\% & 0.3229 & 17.2 & Intermediate operating point \\
        10\% & 0.4916 & 21.9 & Higher detection with more alerts \\
        \bottomrule
    \end{tabular}
\end{table*}

\section{Protocol deviations and reporting boundaries}

An implementation error in the independent Student-$t$ baseline caused a broadcasting failure before any metric entered a result table. We quarantined the failed runs, corrected the likelihood, and reran all five affected seeds under the same protocol. Some legacy core checkpoints lacked paper-ready prediction artifacts. We verified their model, seed, split, and data identities, then reloaded and evaluated them without retraining.

Two boundaries remain. First, no frozen time-varying treatment table is available, so treatment-input performance does not support a contribution. Second, the endpoint lacks independently reconstructed infection evidence and the full SOFA variables. We therefore call the risk outcome dataset-defined terminal positivity, not sepsis probability. These restrictions do not invalidate the six-variable trajectory evaluation, but they limit interpretation of the secondary branch.

\newpage
\section*{NeurIPS Paper Checklist}

\begin{enumerate}

\item {\bf Claims}
    \item[] Question: Do the main claims made in the abstract and introduction accurately reflect the paper's contributions and scope?
    \item[] Answer: \answerYes{}
    \item[] Justification: The abstract, Introduction, Results, and Discussion distinguish competitive point forecasting and stronger probabilistic scores from the weaker secondary risk result.

\item {\bf Limitations}
    \item[] Question: Does the paper discuss the limitations of the work performed by the authors?
    \item[] Answer: \answerYes{}
    \item[] Justification: The dedicated Limitations, ethics, and broader impact section discusses the single-center setting, restricted variables, endpoint validity, subgroup size, calibration, and deployment risks.

\item {\bf Theory assumptions and proofs}
    \item[] Question: For each theoretical result, does the paper provide the full set of assumptions and a complete (and correct) proof?
    \item[] Answer: \answerNA{}
    \item[] Justification: The paper presents an empirical forecasting method and does not claim new theoretical results or formal proofs.

    \item {\bf Experimental result reproducibility}
    \item[] Question: Does the paper fully disclose all the information needed to reproduce the main experimental results of the paper to the extent that it affects the main claims and/or conclusions of the paper (regardless of whether the code and data are provided or not)?
    \item[] Answer: \answerYes{}
    \item[] Justification: Sections 4--6 and Appendix A specify cohort construction, splits, variables, optimization, seeds, metrics, baselines, ablations, and patient-level bootstrap procedures.

\item {\bf Open access to data and code}
    \item[] Question: Does the paper provide open access to the data and code, with sufficient instructions to faithfully reproduce the main experimental results, as described in supplemental material?
    \item[] Answer: \answerNo{}
    \item[] Justification: The current manuscript package does not include a public code repository or redistributable derivative data. MIMIC-IV access remains governed by PhysioNet credentialing and data-use requirements.

\item {\bf Experimental setting/details}
    \item[] Question: Does the paper specify all the training and test details (e.g., data splits, hyperparameters, how they were chosen, type of optimizer) necessary to understand the results?
    \item[] Answer: \answerYes{}
    \item[] Justification: Section 6 reports the fixed split, targets, optimizer, learning rate, weight decay, gradient clipping, batch size, maximum epochs, early stopping, seeds, and evaluation metrics.

\item {\bf Experiment statistical significance}
    \item[] Question: Does the paper report error bars suitably and correctly defined or other appropriate information about the statistical significance of the experiments?
    \item[] Answer: \answerYes{}
    \item[] Justification: Primary results report five-seed mean and standard deviation; paired comparisons additionally use 2,000 patient-level bootstrap replicates per seed, with details in Section 7 and Appendix A.

\item {\bf Experiments compute resources}
    \item[] Question: For each experiment, does the paper provide sufficient information on the computer resources (type of compute workers, memory, time of execution) needed to reproduce the experiments?
    \item[] Answer: \answerNo{}
    \item[] Justification: Parameter counts, training times, memory, latency, and throughput are reported, but the exact GPU model and total project compute were not preserved in the final portable result bundle.
    
\item {\bf Code of ethics}
    \item[] Question: Does the research conducted in the paper conform, in every respect, with the NeurIPS Code of Ethics \url{https://neurips.cc/public/EthicsGuidelines}?
    \item[] Answer: \answerYes{}
    \item[] Justification: The study uses deidentified retrospective data, limits clinical claims, reports negative findings, and discusses misuse, subgroup, privacy, and deployment risks.

\item {\bf Broader impacts}
    \item[] Question: Does the paper discuss both potential positive societal impacts and negative societal impacts of the work performed?
    \item[] Answer: \answerYes{}
    \item[] Justification: The Limitations, ethics, and broader impact section discusses inspectable forecasting as a potential benefit and false reassurance, alert fatigue, subgroup heterogeneity, and workflow shift as potential harms.
    
\item {\bf Safeguards}
    \item[] Question: Does the paper describe safeguards that have been put in place for responsible release of data or models that have a high risk for misuse (e.g., pre-trained language models, image generators, or scraped datasets)?
    \item[] Answer: \answerNA{}
    \item[] Justification: The paper does not release a high-risk foundation model or a newly scraped dataset; patient-level MIMIC-IV-derived records are not redistributed.

\item {\bf Licenses for existing assets}
    \item[] Question: Are the creators or original owners of assets (e.g., code, data, models), used in the paper, properly credited and are the license and terms of use explicitly mentioned and properly respected?
    \item[] Answer: \answerNo{}
    \item[] Justification: MIMIC-IV is cited and its access conditions are acknowledged, but the exact source-version and license statement for the locally prepared derivative were not preserved in the portable experiment bundle.

\item {\bf New assets}
    \item[] Question: Are new assets introduced in the paper well documented and is the documentation provided alongside the assets?
    \item[] Answer: \answerNA{}
    \item[] Justification: No new public dataset, pretrained model, or benchmark asset is released with this manuscript version.

\item {\bf Crowdsourcing and research with human subjects}
    \item[] Question: For crowdsourcing experiments and research with human subjects, does the paper include the full text of instructions given to participants and screenshots, if applicable, as well as details about compensation (if any)? 
    \item[] Answer: \answerNA{}
    \item[] Justification: The work is a retrospective analysis of an existing deidentified critical-care database and involves no crowdsourcing or prospective participant interaction.

\item {\bf Institutional review board (IRB) approvals or equivalent for research with human subjects}
    \item[] Question: Does the paper describe potential risks incurred by study participants, whether such risks were disclosed to the subjects, and whether Institutional Review Board (IRB) approvals (or an equivalent approval/review based on the requirements of your country or institution) were obtained?
    \item[] Answer: \answerNA{}
    \item[] Justification: No new human-subject recruitment or intervention was conducted; the study uses deidentified MIMIC-IV-derived data under the database access requirements.

\item {\bf Declaration of LLM usage}
    \item[] Question: Does the paper describe the usage of LLMs if it is an important, original, or non-standard component of the core methods in this research? Note that if the LLM is used only for writing, editing, or formatting purposes and does \emph{not} impact the core methodology, scientific rigor, or originality of the research, declaration is not required.
    \item[] Answer: \answerNA{}
    \item[] Justification: LLMs are not an important, original, or non-standard component of the model, data construction, or experimental methodology.

\end{enumerate}

\end{document}